\pdfoutput=1
\documentclass[10pt]{article}
\usepackage[preprint]{tmlr}
\usepackage[T1]{fontenc}
\usepackage{amsmath,amssymb,amsthm,mathtools}
\usepackage{graphicx,booktabs,array,microtype,xspace}
\usepackage{hyperref,url}
\hypersetup{colorlinks=true,citecolor=blue,linkcolor=blue,urlcolor=blue}
\newtheorem{theorem}{Theorem}[section]

\newtheorem{corollary}[theorem]{Corollary}
\newtheorem{assumption}[theorem]{Assumption}

\newcommand{\R}{\mathbb R}
\newcommand{\E}{\mathbb E}
\newcommand{\U}{\mathcal U}
\newcommand{\N}{\mathcal N}
\newcommand{\KL}{\operatorname{KL}}

\newcommand{\TV}{\operatorname{TV}}
\newcommand{\norm}[1]{\left\lVert #1\right\rVert}
\newcommand{\method}{NormFlowSDE\xspace}
\title{NormFlowSDE: Normalizing Flows for\\Amortized Bayesian State Estimation}
\author{\name Yi Zhang \email yi.zhang@uni-a.de \\
        \name Lars Mikelsons \email lars.mikelsons@uni-a.de \\
        \addr Chair of Mechatronics, University of Augsburg, Germany}
\begin{document}
\maketitle
\begin{abstract}
Bayesian state estimators, such as Kalman filters, estimate states from noisy observations based on dynamical models. However, obtaining accurate models can be challenging when the system involves effects that are difficult to model faithfully, such as friction, impacts, and deformation in contact-rich manipulation. Uncertainty in system parameters, inputs, and disturbances, together with simplified noise models, can further introduce simulation-to-reality mismatch and degrade estimation accuracy. These limitations motivate learning a distribution over state trajectories conditioned on observations and controls without explicitly specifying the underlying physical models.
We introduce NormFlowSDE, an amortized Bayesian state estimator that learns a distribution over state trajectories by combining neural stochastic differential equations with conditional normalizing flows.
Its key design choice is to utilize the capability of normalizing flows to learn stochastic behavior of variables, while leveraging the continuous-time dynamics of SDEs to model the evolution of states over time.
Once trained, the model draws state paths for new measurement sequences without per-sequence optimization. 
In addition, we connect NormFlowSDE to stochastic continuous-time dynamics by proving weak convergence of a mean-centered, drift-corrected inference scheme to an It\^o diffusion under stated assumptions.
Finally, we benchmark NormFlowSDE against direct estimators and learned particle filters based on normalizing flows on partially observed dynamical systems, including simulated systems and real-world robot datasets.
The results show that NormFlowSDE attains the lowest estimation error with calibrated uncertainty intervals on both robot datasets and filters four times faster than the fastest particle filter on every dataset.
\end{abstract}
\section{Introduction}\label{sec:introduction}
Bayesian inference provides a principled framework for estimating unknown quantities from noisy observations. Conditional normalizing flows provide a way towards amortized Bayesian inference, which means making Bayesian parameter identification reusable across datasets \citep{radev2021amortized}. Trained on simulated parameter--observation pairs, BayesFlow \citep{radev2020bayesflow} combines a learned summary of the observations with an invertible transformation to approximate the parameter posterior. Once fitted, the model produces posterior samples for new observations without repeating parameter optimization. This amortization makes expressive Bayesian inference practical for repeated identification tasks. Its success motivates the question studied here: how can conditional normalizing flows be combined with temporal dynamics to infer evolving states and their uncertainty?

State estimation extends the inference target from a fixed parameter vector to a sequence of unknown states. Noisy or partial measurements constrain that sequence, while temporal dependence links uncertainty at one time to states at subsequent times. A useful estimator must therefore combine flexible conditional distributions, information from the observations, and a model of temporal evolution. Stochastic differential equations (SDEs) provide a natural starting point. In its standard form,
\begin{equation}\label{eq:sde-standard}
 dz_t=f(z_t,t)\,dt+\sigma(z_t,t)\,dW_t,
\end{equation}
an SDE combines a drift $f$, which describes the nominal evolution of the state $z_t$, with a diffusion $\sigma$ that scales the increments of a Wiener process $W_t$ and represents variability around it. A neural SDE parameterizes $f$ and $\sigma$ by neural networks, which can be trained in two established ways: Latent SDE and SDE-GAN.

Latent SDEs \citep{li2020scalable} represent time series through an underlying stochastic process and learn its dynamics through variational inference, coupling a generative SDE with an approximate posterior process. When the observation model uses a Gaussian decoder with learned variance, it can inherit a decoder variance-collapse problem. The estimated variance may approach zero, which makes training highly sensitive to small reconstruction errors and destabilizes optimization \citep{takahashi2018student,rybkin2021calibrated}. In addition, every increment of the posterior process is Gaussian given the current state, so a non-Gaussian conditional distribution of the next state can only arise through the nonlinearity of drift and diffusion over several steps.

SDE-GANs \citep{kidger2021gan} instead treat a neural SDE as a trajectory generator and train it against a discriminator to match observed path distributions. This framework accommodates flexible stochastic dynamics, but it requires coupled adversarial optimization and careful discriminator regularization \citep{kidger2021gan}. Moreover, it does not provide a tractable finite-interval transition likelihood, which restricts state estimation to simulation-based bootstrap filtering and rules out observation-informed proposals and likelihood evaluation between measurement times. Signature-kernel scores remove the discriminator and train the same generator with a proper scoring rule on paths \citep{issa2023sigkernel}.

\begin{figure}[htbp]
\centering
\includegraphics[width=\linewidth]{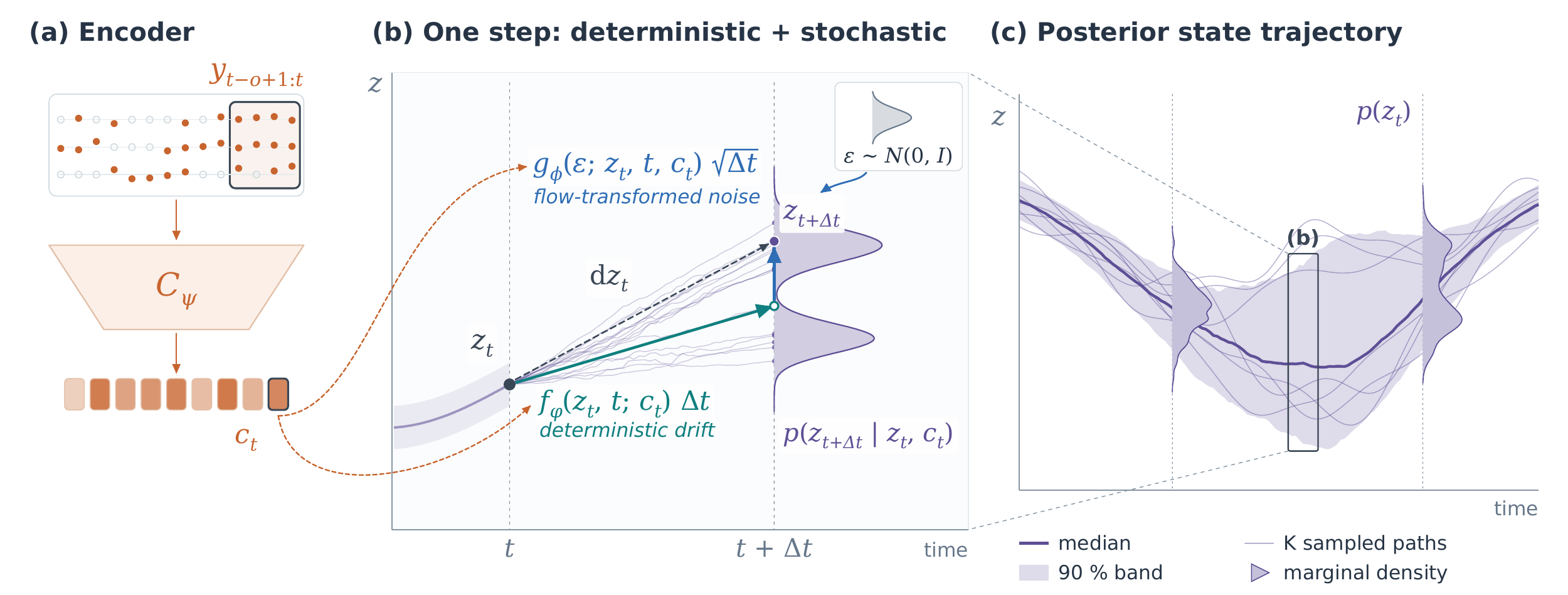}
\caption{NormFlowSDE estimates states conditional on observations. (a) At each step, the encoder $C_\psi$ maps the trailing window of the last $o$ measurements $y_{t-o+1:t}$ (boxed), which ends before the state being estimated, to a context vector $c_t$ (orange), which conditions both the drift and the noise model. (b) Each state update combines a deterministic drift increment (green) with flow-transformed Gaussian noise (blue), yielding a flexible conditional distribution for the next state $z_{t+\Delta t}$ (purple). (c) Repeated updates from the supplied initial state generate $K$ sampled state trajectories. The plot shows their pointwise median, 90\% uncertainty bands, and marginal state densities at selected times. 
The label $p(z_t)$ is shorthand for the marginal density of $z_t$ conditional on the observation windows and controls, which are held fixed across samples. The boxed interval is expanded in (b). All curves and densities are schematic.}
\label{fig:overview}
\end{figure}

In contrast to these two neural SDE methods, we seek to combine the temporal structure of neural SDEs with the explicit conditional densities and direct sampling offered by normalizing flows (NFs) \citep{dinh2014nice,rezende2015flows,dinh2017realnvp}. The flow transforms the noise within each state update while preserving tractable likelihoods and the sequential structure of neural SDEs. We call this construction NormFlowSDE. An observation-conditioned neural drift supplies a nominal state update, and an invertible conditional flow transforms Gaussian inputs into state-dependent noise. Both are trained jointly from paired state--measurement data. Figure~\ref{fig:overview} shows the resulting estimator.

The construction also raises a mathematical question: under what conditions does the flow-based state update admit a continuous-time SDE limit? In Appendix~\ref{app:diffusion}, we prove that, with the learned coefficients fixed and under explicit regularity and moment assumptions, the trajectories of a centered inference scheme converge in distribution to the unique solution of an It\^o SDE as the time step tends to zero. 

Section~\ref{sec:experiments} evaluates NormFlowSDE as a causal amortized filter on two simulated systems (a controlled pendulum and the Lorenz--96 model) and two robot datasets (ALOHA insertion \citep{zhao2023aloha} and REASSEMBLE \citep{sliwowski2025reassemble}). In addition, we implement NormFlowSDE as a transition model in a particle filter. Both implementations are thoroughly benchmarked. First, NormFlowSDE as a direct amortized filter is compared with latent Neural Stochastic Flow (latent NSF), latent SDE, and a neural SDE trained with the signature-kernel score (SigKer SDE). Second, NormFlowSDE in a particle filter is compared with three NF-based methods: an NF-based differentiable particle filter (NF-DPF), Neural Stochastic Flow (NSF), and a flow-matching-based SDE (FlowSDE). The benchmark methods are described in detail in Section~\ref{sec:related}. 

Several other works use generative methods to model stochastic processes directly. The continuous-time flow process (CTFP) \citep{deng2020ctfp} maps a Wiener process through a time-indexed invertible transformation, which gives the observed process tractable finite-dimensional likelihoods. Continuous latent process flows \citep{deng2021clpf} apply such a transformation to a latent process driven by an SDE. The GP-VAE \citep{fortuin2020gpvae} combines Gaussian-process priors with deep encoders, and neural processes learn distributions over functions \citep{garnelo2018neural}. Diffusion and flow-matching models have also been applied directly to time series. TSDiff \citep{kollovieh2023tsdiff} trains an unconditional diffusion model over whole windows and obtains forecasts by guiding it toward the observed past at sampling time. TSFlow \citep{kollovieh2025tsflow} uses conditional flow matching with Gaussian-process priors to generate and forecast series. Stochastic process diffusion generates whole series as continuous functions \citep{bilos2023stochastic}. All of these models target the distribution of the entire time series and do not model a process in physical time. They therefore provide neither a transition model for a recursive filter nor a causal state estimator comparable with the methods in our work, and we do not include them in the benchmark.

\section{Related Work}\label{sec:related}
For Bayesian state estimation, the relevant question is how a learned model combines temporal dynamics and measurements to represent uncertainty over the states. Learned estimators take one of two routes. A \emph{direct estimator} learns the conditional distribution of the state trajectory given the measurements, which enter the estimator through an encoder. Direct estimators include NormFlowSDE (Section~\ref{sec:method}), latent NSF (Section~\ref{sec:related-latent-nsf}), latent SDE (Section~\ref{sec:related-latent-sde}), and SigKer SDE (Section~\ref{sec:related-sigker}). A \emph{particle-filter estimator} learns a transition model and a measurement model separately and combines them by Bayesian weighting, so that the measurements enter only through the likelihood of the measurement model. Particle-filter estimators include NF-DPF (Section~\ref{sec:related-nfdpf}), NSF-PF (Section~\ref{sec:related-latent-nsf}), FlowSDE-PF (Section~\ref{sec:related-flowsde}), and NormFlowSDE-PF, which is included for completeness. Throughout, $x_k$ denotes the state, $y_k$ the measurement, and $u_k$ the control at step $k$.

\subsection{Direct Estimators}\label{sec:related-direct}
Conditional normalizing flows provide a route to Bayesian inference by learning a distribution over unknown quantities conditioned on observed data. For an unknown parameter vector $\vartheta$ and observations $y$, an invertible map $\vartheta=g_\phi(\epsilon;y)$ transforms a simple base variable $\epsilon\sim\N(0,I)$ into samples from a normalized conditional density $q_\phi(\vartheta\mid y)$, evaluated through change of variables \citep{dinh2014nice,rezende2015flows}. The invertible maps are compositions $g_\phi=f_{\phi,L_{\mathrm f}}\circ\cdots\circ f_{\phi,1}$ of $L_{\mathrm f}$ affine coupling layers \citep{dinh2017realnvp}. Each layer splits its input $\zeta$ into coordinate blocks $(\zeta_A,\zeta_B)$ and, given a conditioning input $c$, produces
\begin{equation}\label{eq:affine-coupling}
 \zeta'_A=\zeta_A,\qquad
 \zeta'_B=\zeta_B\odot\exp\bigl(\kappa_\ell(\zeta_A,c)\bigr)+\beta_\ell(\zeta_A,c),
\end{equation}
where $\odot$ denotes componentwise multiplication. The networks $\kappa_\ell$ and $\beta_\ell$, whose parameters are included in $\phi$, output the log scales and shifts.

Given training pairs from $p(\vartheta,y)=p(\vartheta)p(y\mid\vartheta)$, conditional maximum likelihood minimizes the expected forward KL divergence from the target posterior to its learned approximation \citep{radev2020bayesflow}:
\begin{equation}\label{eq:amortized-forward-kl}
 \E_{p(\vartheta,y)}[-\log q_\phi(\vartheta\mid y)]
 =\E_{p(y)}\,\left[\KL\,\left(p(\vartheta\mid y)\,\Vert\,q_\phi(\vartheta\mid y)\right)\right]+C_0,
\end{equation}
where $C_0$ is the conditional entropy of $\vartheta$ given the observations, independent of $\phi$. Thus, paired simulations supply a posterior-learning objective without requiring pointwise evaluation of the target posterior. 

BayesFlow applies this principle to simulation-based parameter inference, combining a learned summary of the observations with a conditional invertible network \citep{radev2020bayesflow}. Once trained, it produces posterior samples for new datasets generated by the same simulator without per-dataset optimization. Its target is a fixed parameter vector, so the construction must be modified before it can infer an evolving state. \citet{rasul2021conditionedflows} condition normalizing flows on recurrent or attention-based representations of past observations and covariates and generate multivariate forecast trajectories autoregressively. In this formulation the flow learns a joint distribution of future states, current states, and past observations, which makes learning difficult and limits the forecast horizon.

The direct estimators in our benchmark share one interface. A causal encoder summarizes the measurements up to step $k$ in a context $c_k$, and a stochastic update produces $x_{k+1}$ from $x_k$ and $c_k$, starting from the known initial state. Repeated updates give $M$ sampled trajectories. The four estimators differ in the distribution of this update and in how it is trained. NormFlowSDE (Section~\ref{sec:method}) uses an observation-conditioned drift and a conditional normalizing flow on the noise of each update, trained on full rollouts. Latent NSF keeps the normalizing flow in the prior and estimates with a Gaussian posterior (Section~\ref{sec:related-latent-nsf}). The latent SDE (Section~\ref{sec:related-latent-sde}) and the signature-kernel SDE (Section~\ref{sec:related-sigker}) use Gaussian increments with a learned diffusion and are trained by a variational bound and by a proper scoring rule on paths, respectively.

\subsubsection{Neural Stochastic Flows and Latent NSF}\label{sec:related-latent-nsf}
Neural Stochastic Flows (NSFs) use a conditional normalizing flow to model the distribution of a future state $x_t$ given a starting state $x_s$ and a time interval $h=t-s>0$ \citep{kiyohara2025nsf}. The sampling procedure has two stages: a neural network constructs a Gaussian candidate $\widetilde x$, and an invertible network $g_\phi$ transforms that candidate into $x_t$. The context is $(x_s,h,s)$, with absolute time $s$ omitted for autonomous dynamics. The two sampling stages are
\begin{equation}\label{eq:nsf-transition-sampling}
\begin{aligned}
 \widetilde x&=x_s+h\mu_\theta(x_s,h,s)+\sqrt h\,\sigma_\theta(x_s,h,s)\odot\epsilon,
 \qquad \epsilon\sim\N(0,I),\\
 x_t&=g_\phi(\widetilde x;x_s,h,s).
\end{aligned}
\end{equation}
Here $\mu_\theta$ and $\sigma_\theta>0$ are outputs of ordinary multilayer perceptrons that set the Gaussian displacement and noise scale. The factors $h$ and $\sqrt h$ are written out although $h$ is also an input, so that $\widetilde x\to x_s$ as $h\to0$ whatever the networks output. The flow $g_\phi$ is a composition of affine coupling layers (Equation~\eqref{eq:affine-coupling}) conditioned on $(x_s,h,s)$, whose log scales and shifts are additionally multiplied by $h$, in contrast to Equation~\eqref{eq:affine-coupling}:
\begin{equation}\label{eq:nsf-affine-coupling}
 \zeta'_B=\zeta_B\odot\exp\bigl(h\,\kappa_\ell(\zeta_A,x_s,h,s)\bigr)+h\,\beta_\ell(\zeta_A,x_s,h,s).
\end{equation}
For $h>0$ the inverse transformation and its Jacobian give an evaluable transition density. Sampling evaluates Equation~\eqref{eq:nsf-transition-sampling} once for the whole gap $h$, and the Euler--Maruyama-type candidate is formed in a single step without SDE integration.

Transition densities of a Markov process satisfy the Chapman--Kolmogorov relation: a direct transition must agree with two successive transitions after averaging over the intermediate state \citep[Chapter~4, Section~1]{ethier1986markov}. NSF encourages its learned densities to satisfy this relation,
\begin{equation}\label{eq:nsf-chapman-kolmogorov}
 p_\omega^{s,t}(x_t\mid x_s)
 \approx\int p_\omega^{r,t}(x_t\mid x_r)\,
               p_\omega^{s,r}(x_r\mid x_s)\,dx_r,
 \qquad s<r<t,
\end{equation}
where the approximation sign indicates that consistency is encouraged rather than guaranteed. The integral is generally intractable, so NSF introduces an auxiliary bridge distribution $q_\xi^{\mathrm{br}}(x_r\mid x_s,x_t;s,r,t)$ with parameters $\xi$, which yields variational upper bounds on the KL divergences between the direct and composed transitions in both directions. Averaging these bounds over sampled time triplets defines the consistency penalty $\mathcal L_{\mathrm{flow}}(\omega,\xi)$, and training on observed state pairs $\mathcal D$ minimizes
\begin{equation}\label{eq:nsf-training-objective}
 \mathcal L(\omega,\xi)
 =-\E_{(x_s,x_t,s,t)\sim\mathcal D}
       \log p_\omega^{s,t}(x_t\mid x_s)
   +\lambda\mathcal L_{\mathrm{flow}}(\omega,\xi),
 \qquad \lambda\geq0.
\end{equation}

The NSF transition is a model of the dynamics. The published route from NSF to state inference is latent NSF \citep[Section~4]{kiyohara2025nsf}, a variational state-space model whose prior transition is an NSF and whose observations follow an emission model $p_\psi(y_k\mid x_k)$. Its inference network is a recurrent encoder of the past observations that outputs a Gaussian with diagonal covariance for each time,
\begin{equation}\label{eq:latent-nsf-posterior}
 q_\varphi(x_{t_i}\mid y_{\le t_i})=\N\,\left(x_{t_i};\,m_{t_i},\,\mathrm{diag}\,s_{t_i}^2\right),
 \qquad (m_{t_i},s_{t_i})=\mathrm{GRU}_\varphi(y_{t_i},\Delta t_i,h_{t_{i-1}}).
\end{equation}
Training minimizes a $\beta$-weighted negative evidence lower bound, which compares each posterior marginal with the NSF prediction from the previous posterior, together with the flow consistency penalty. In our adaptation the posterior is additionally conditioned on the previous sampled state and rolled out from the known initial state (Appendix~\ref{app:direct-baselines}).

\subsubsection{Latent SDEs}\label{sec:related-latent-sde}
Latent SDEs \citep{li2020scalable} model a time series through a prior SDE and an approximate posterior SDE that share their diffusion,
\begin{equation}\label{eq:latent-sde}
 \text{prior:}\quad dx=f_\theta(x,t)\,dt+g_\theta(x,t)\,dW_t,
 \qquad
 \text{posterior:}\quad dx=h_\varphi(x,c_t,t)\,dt+g_\theta(x,t)\,dW_t,
\end{equation}
where the posterior drift $h_\varphi$ also receives an encoding $c_t$ of the observations. Because the diffusions agree, Girsanov's theorem \citep[Sec.~8.6]{oksendal2003sde} makes the KL divergence between the two path measures finite \citep{li2020scalable},
\begin{equation}\label{eq:latent-sde-kl}
 \KL=\E\left[\int_0^T\tfrac12\|u_t\|^2\,dt\right],
 \qquad
 u_t=\frac{h_\varphi(x_t,c_t,t)-f_\theta(x_t,t)}{g_\theta(x_t,t)},
\end{equation}
where the expectation is over posterior paths. Training maximizes the evidence lower bound, the expected log likelihood of the observations under a decoder $p(y_{t_i}\mid x_{t_i})$ minus this path KL, with gradients through the SDE solver obtained by a stochastic adjoint. With a Gaussian decoder of learned variance, this objective can drive the variance towards zero. Since the posterior drift is conditioned on the encoded observations, posterior paths sampled from the initial state are observation-conditioned state trajectories. A latent SDE with a causal encoder is therefore a direct estimator. Its increments are Gaussian given the current state, so non-Gaussian marginals arise only from the nonlinearity of drift and diffusion.

\subsubsection{Neural SDEs Trained with Signature-Kernel Scores}\label{sec:related-sigker}
SDE-GANs \citep{kidger2021gan} train a neural SDE generator $dx=\mu_\theta(x,t)\,dt+\sigma_\theta(x,t)\,dW_t$ adversarially against a path discriminator. \citet{issa2023sigkernel} replace the discriminator by a scoring rule on path space. For a model path law $P$ and a realized path $x$, the signature-kernel score is
\begin{equation}\label{eq:sigker-score}
 \mathcal S_k(P,x)=\E_{X,X'\sim P}\bigl[k(X,X')\bigr]-2\,\E_{X\sim P}\bigl[k(X,x)\bigr],
\end{equation}
where $X$ and $X'$ are independent generator paths and $k$ is the signature kernel. This kernel is the inner product of the path signatures after the paths are lifted with a static kernel, such as an RBF kernel. It is computed for a pair of discretized paths by solving a Goursat partial differential equation \citep{salvi2021goursat}. With a characteristic kernel the score is strictly proper, so its expectation over data is minimized only by the data-generating path law. Up to a term independent of $P$, it equals the squared maximum mean discrepancy. Given $m$ generator paths, both expectations have unbiased Monte Carlo estimates, and training needs neither a discriminator nor a min--max optimization. When the drift and diffusion also receive an encoding of the measurements and the score compares the generated paths with the paired reference states, the generator becomes a direct estimator of the state trajectory. As for the latent SDE, its increments are Gaussian given the state. However, it has no tractable transition density, since the score is evaluated only from samples.

\subsection{Particle-Filter Estimators}\label{sec:related-pf}
Bayesian filtering updates the distribution of a hidden state as new measurements arrive. A particle filter approximates this distribution, $p(x_k\mid y_{0:k})$, using $M$ candidate states, called particles, and weights describing their relative contributions. The bootstrap particle filter first propagates each previous particle through the transition model by drawing $x_k^{(i)}\sim p(\cdot\mid x_{k-1}^{(i)})$ \citep{gordon1993bootstrap}. The observation likelihood $p(y_k\mid x_k^{(i)})$ then measures how well each predicted state explains the measurement $y_k$, and the normalized weights are
\begin{equation}\label{eq:bootstrap-weights}
 w_k^{(i)}=
 \frac{w_{k-1}^{(i)}p(y_k\mid x_k^{(i)})}
 {\sum_{j=1}^{M}w_{k-1}^{(j)}p(y_k\mid x_k^{(j)})}.
\end{equation}
The numerator combines information from earlier measurements, carried by $w_{k-1}^{(i)}$, with the new evidence. A resampling step then draws $M$ particles with replacement according to these weights and resets them to $1/M$. Some particle filters instead sample from a proposal $q$ that incorporates the new measurement, such as Gaussian proposals from local linearization \citep{doucet2000sequential} or the unscented particle filter \citep{vandermerwe2000unscented}. Each weight is then also multiplied by the ratio of transition and proposal densities, which equals one for the bootstrap filter. Exact filtering without particles is possible when the latent dynamics are restricted, as in Normalizing Kalman Filters, which combine an invertible observation transform with a linear Gaussian latent model \citep{debezenac2020nkf}. However, observation-likelihood training alone does not tie their latent coordinates to the physical states, so the filtered latent state is not an estimate of the physical state.

A learned particle filter needs a transition model that can be sampled and a measurement model whose density can be evaluated. NF-DPF learns its transition, proposal, and measurement distributions with normalizing flows by design, and its proposal is conditioned on the current measurement. For comparability in our benchmark, NormFlowSDE-PF, NSF-PF, and FlowSDE-PF are bootstrap filters. Their proposal is the learned transition itself, which is conditioned on the control and the time but not on the measurement, so the measurement $y_k$ enters only through the weights of Equation~\eqref{eq:bootstrap-weights}. Their measurement density $q_\psi(y_k\mid x_k)$ is a conditional normalizing flow with the same architecture and the same maximum-likelihood training in all three filters. Each filter fits its own instance jointly with its transition, so no trained network is shared between filters. Thus, they differ only in the transition model.

\subsubsection{Normalizing Flow-Based Differentiable Particle Filters}\label{sec:related-nfdpf}
\citet{chen2025flowparticle} introduce NF-DPF, which learns the transition, proposal, and measurement distributions using three invertible maps. Subscripts $\mathrm{dyn}$, $\mathrm{prop}$, and $\mathrm{obs}$ distinguish their separate parameter sets. With tractable base densities $g_{\theta_{\mathrm{dyn}}}$ and $g_{\theta_{\mathrm{prop}}}$, these models take the form
\begin{equation}\label{eq:nfdpf-components}
\begin{alignedat}{3}
 \mathrm{Transition}&:\quad& x_k^{\mathrm{dyn}}&=f_{\phi_{\mathrm{dyn}}}(\zeta_k^{\mathrm{dyn}}),
 &\qquad\zeta_k^{\mathrm{dyn}}&\sim g_{\theta_{\mathrm{dyn}}}(\cdot\mid x_{k-1}),\\
 \mathrm{Proposal}&:\quad& x_k^{\mathrm{prop}}&=f_{\phi_{\mathrm{prop}}}(\zeta_k^{\mathrm{prop}};y_k),
 &\qquad\zeta_k^{\mathrm{prop}}&\sim g_{\theta_{\mathrm{prop}}}(\cdot\mid x_{k-1},y_k),\\
 \mathrm{Measurement}&:\quad& y_k&=f_{\phi_{\mathrm{obs}}}(\epsilon_k;x_k),
 &\qquad\epsilon_k&\sim\N(0,I).
\end{alignedat}
\end{equation}
The parameter vectors $\omega_{\mathrm{dyn}}=(\theta_{\mathrm{dyn}},\phi_{\mathrm{dyn}})$ and $\omega_{\mathrm{prop}}=(\theta_{\mathrm{prop}},\phi_{\mathrm{prop}})$ collect the base-density and flow parameters. The first map defines the transition density $p_{\omega_{\mathrm{dyn}}}(x_k\mid x_{k-1})$, the second the proposal $q_{\omega_{\mathrm{prop}}}(x_k\mid x_{k-1},y_k)$, from which the particles are sampled, and the third describes possible observations given a state. Invertibility makes all three densities evaluable by change of variables. For example, setting $\epsilon_k=f_{\phi_{\mathrm{obs}}}^{-1}(y_k;x_k)$ gives the measurement likelihood
\begin{equation}\label{eq:nfdpf-observation-likelihood}
 p_{\phi_{\mathrm{obs}}}(y_k\mid x_k)
 =p_\epsilon(\epsilon_k)\left|\det\frac{\partial f_{\phi_{\mathrm{obs}}}^{-1}(y_k;x_k)}{\partial y_k}\right|,
\end{equation}
where $p_\epsilon$ is the standard Gaussian density and $x_k$ is held fixed in the Jacobian. For each proposed particle $x_k^{(i)}\sim q_{\omega_{\mathrm{prop}}}(\cdot\mid x_{k-1}^{(i)},y_k)$, these densities enter the update
\begin{equation}\label{eq:nfdpf-weights}
 w_k^{(i)}\propto w_{k-1}^{(i)}
 \frac{\underbrace{p_{\phi_{\mathrm{obs}}}(y_k\mid x_k^{(i)})}_{\mathrm{measurement\ likelihood}}\,
 \underbrace{p_{\omega_{\mathrm{dyn}}}(x_k^{(i)}\mid x_{k-1}^{(i)})}_{\mathrm{transition\ density}}}
 {\underbrace{q_{\omega_{\mathrm{prop}}}(x_k^{(i)}\mid x_{k-1}^{(i)},y_k)}_{\mathrm{proposal\ density}}},
\end{equation}
followed by normalization as in Equation~\eqref{eq:bootstrap-weights}. NF-DPF uses entropy-regularized optimal-transport resampling to support gradient-based learning, and training maximizes a particle estimate of the observation log likelihood or minimizes a supervised state-estimation error. Although the proposal is conditioned on the current measurement, it is trained only as one factor of the importance weight in Equation~\eqref{eq:nfdpf-weights}, so its samples approximate the filtering distribution only after weighting by the transition and measurement densities and resampling. 

\subsubsection{FlowSDE-PF}\label{sec:related-flowsde}
FlowSDE uses conditional flow matching to train a neural vector field from state-transition data, together with a potential landscape for interpreting critical transitions \citep{lou2026flowsde}. The learned field is the drift of the SDE
\begin{equation}\label{eq:flowsde-related-dynamics}
 dX_t=f_\varphi(X_t,t)\,dt+\sqrt{2D}\,dW_t,
\end{equation}
where $W_t$ is standard Brownian motion and $D>0$ is a prescribed scalar diffusion coefficient \citep[Section~1.1]{lou2026flowsde_supp}. The flow-matching step constructs supervision for $f_\varphi$ between observed states. For a pair $(x_s,x_t)$ with gap $h=t-s$, it chooses an intermediate fraction $\tau\in(0,1)$ and draws a state at physical time $r=s+\tau h$ from the Gaussian bridge between the two observed states:
\begin{equation}\label{eq:flowsde-bridge}
 \widetilde X_r=(1-\tau)\,x_s+\tau\,x_t+\sigma_{\mathrm{br}}\sqrt{\tau(1-\tau)}\,\eta,
 \qquad \eta\sim\N(0,I).
\end{equation}
It is the marginal at time $r$ of a Brownian motion with scale $\sigma_{\mathrm{br}}$ pinned to $x_s$ at time $s$ and to $x_t$ at time $t$. Differentiating this bridge in time supplies an analytic vector-field target $b_r^{\mathrm{br}}$, expressed per unit physical time, and the data-fitting loss is
\begin{equation}\label{eq:flowsde-cfm-regression}
 \mathcal L_{\mathrm{CFM}}(\varphi)
 =\E\,\left[
 \left\|f_\varphi(\widetilde X_r,r)
       -b_r^{\mathrm{br}}(\widetilde X_r\mid x_s,x_t)\right\|^2\right],
\end{equation}
which is averaged over training pairs, intermediate times, and bridge samples. Flow matching thus turns dynamics learning into vector-field regression, without simulating trajectories or differentiating through a solver during training. A separate network represents a scalar potential $V_\eta$. It has no target of its own and is trained jointly with the drift through the regularizers of Equation~\eqref{eq:flowsde-training-objective} below, which make $-\nabla V_\eta$ the gradient part of the drift in the decomposition
\begin{equation}\label{eq:flowsde-potential-decomposition}
 f_\varphi(x,t)=-\nabla V_\eta(x)+R_{\varphi,\eta}(x,t),
 \qquad R_{\varphi,\eta}=f_\varphi+\nabla V_\eta,
\end{equation}
with a gradient component towards lower potential and a non-conservative residual force. The full objective,
\begin{equation}\label{eq:flowsde-training-objective}
 \mathcal L(\varphi,\eta)
 =\mathcal L_{\mathrm{CFM}}(\varphi)
 +\lambda_{\mathrm E}\mathcal L_{\mathrm{EPR}}(\varphi,\eta)
 +\lambda_{\mathrm H}\mathcal L_{\mathrm{HJB}}(\varphi,\eta)
 +\lambda_{\mathrm O}\mathcal L_{\mathrm{orth}}(\varphi,\eta),
\end{equation}
adds soft physical-consistency penalties with nonnegative weights: an entropy-production penalty on the residual force under weights derived from $\exp(-V_\eta/D)$, a Hamilton--Jacobi--Bellman penalty that encourages compatibility between the drift and the stationary density $p_{\mathrm{ss}}\propto\exp(-V_\eta/D)$, and a penalty on the alignment of $R_{\varphi,\eta}$ with $\nabla V_\eta$ \citep[Sections~1.2--1.4]{lou2026flowsde_supp}.

FlowSDE has no inference component. Its flow-matching targets are bridges between consecutive states, its uncertainty is fixed by the prescribed $D$, and its potential regularizers assume a stationary density of the autonomous dynamics. Conditioning the drift on measurements would leave the uncertainty fixed by $D$ instead of learning it from the observations, and would remove the interpretation of the potential. In FlowSDE-PF the drift is therefore conditioned on the state, the time, and the control, and particles are propagated by Euler--Maruyama integration of Equation~\eqref{eq:flowsde-related-dynamics}.

\section{NormFlowSDE for Amortized Bayesian State Estimation}\label{sec:method}

\subsection{Observation-Conditioned State Transitions}

To extend conditional normalizing flows to amortized Bayesian state estimation, we represent the state as a time-dependent random variable and combine a conditional flow with the temporal update structure of a neural SDE. With control inputs, the neural SDE of Equation~\eqref{eq:sde-standard} reads
\begin{equation}\label{eq:neural-sde}
 dz_t=f_\varphi(z_t,t,u_t)\,dt+\sigma(z_t,t,u_t)\,dW_t,
\end{equation}
where $z_t\in\R^d$ is the state variable, $u_t$ denotes control inputs, and $W_t$ is a $d$-dimensional Wiener process. The drift $f_\varphi$ determines the infinitesimal conditional mean, while $\sigma\sigma^\top$ determines the conditional covariance per unit time. At times $t_k=k\Delta t$, with a fixed interval $\Delta t>0$, the Euler--Maruyama discretization of Equation~\eqref{eq:neural-sde} is
\begin{equation}\label{eq:euler-maruyama}
 z_{k+1}=z_k+\Delta t\,f_\varphi(z_k,t_k,u_k)+\sqrt{\Delta t}\,\sigma(z_k,t_k,u_k)\,\epsilon_k,
 \qquad \epsilon_k\sim\N(0,I_d).
\end{equation}
Its stochastic increment is a linear map of a Gaussian input and therefore Gaussian given $z_k$.

To condition the update on the measurements, an encoder $C_\psi$, the counterpart of the summary network of BayesFlow (Section~\ref{sec:related-direct}), compresses a window of measurements preceding the state being estimated. For an integer window length $o\geq1$, it computes the window feature
\begin{equation}\label{eq:sequence-conditioning}
 c_k=C_\psi\,\left(y_{\max(0,k-o+1):k}\right),
 \qquad k=0,\ldots,N-1.
\end{equation}
Thus the update to $z_{k+1}$ uses at most the last $o$ observations through $t_k$. Near the sequence start, only the available prefix is used. We replace the linear Gaussian noise map $\epsilon_k\mapsto\sigma\epsilon_k$ of Equation~\eqref{eq:euler-maruyama} with a conditional invertible flow to obtain the finite-step NormFlowSDE update
\begin{equation}\label{eq:normflowsde-overview}
 z_{k+1}=z_k+\Delta t\,f_\varphi(z_k,t_k,u_k,c_k)
          +\sqrt {\Delta t}\,g_\phi(\epsilon_k;z_k,t_k,u_k,c_k),
 \qquad \epsilon_k\stackrel{\mathrm{iid}}\sim\N(0,I_d).
\end{equation}
Here $z_k$ denotes a state sample at time $t_k$, and the Gaussian inputs are independent of the supplied data and initial state. The initial state $z_0$ is provided by the data and initializes the stochastic process once. Each subsequent update uses the current generated state $z_k$, time $t_k$, control $u_k$ when available, and window feature $c_k$.

This construction assigns nominal deterministic evolution to the drift network, while the conditional flow learns the distribution of stochastic increments from data. The drift can also incorporate prior knowledge. Where the equations of motion are partly known, $f_\varphi$ can be a hybrid model that adds a learned correction to them. All experiments in this paper nevertheless use a purely learned drift. Both the drift and the flow receive observation information and reuse their fitted parameters across sequences, enabling amortized conditional trajectory sampling. The flow can also contribute to the conditional mean because its output is uncentered. Appendix~\ref{app:diffusion} constructs a centered inference scheme and proves its weak convergence to an It\^o diffusion under stated assumptions.

\paragraph{Shared feature network.}
The drift and the flow are functions of the same inputs $(z_k,t_k,u_k,c_k)$. Instead of processing these inputs in two separate networks, we compute one feature vector per update with a multilayer perceptron $T_\varphi$,
\begin{equation}\label{eq:shared-feature}
 \chi_k=T_\varphi(z_k,t_k,u_k,c_k),
\end{equation}
and let both parts read it. The drift is a linear map of the feature, $f_\varphi(z_k,t_k,u_k,c_k)=A_\varphi\chi_k+b_\varphi$, and the flow depends on its conditioning inputs only through the feature, $g_\phi(\epsilon_k;z_k,t_k,u_k,c_k)=g_\phi(\epsilon_k;\chi_k)$. One update of Equation~\eqref{eq:normflowsde-overview} therefore evaluates $T_\varphi$ once, followed by the linear drift head and the layers of the flow. We keep the notation of Equation~\eqref{eq:normflowsde-overview}. 

We next derive a composite training objective for NormFlowSDE. 

\subsection{Training Objective and Procedure}\label{sec:training}
Training uses paired reference states $x_{0:N}$, observations $y_{0:N-1}$, and controls $u_{0:N-1}$ when available. The reference path $\mathbf x=(x_1,\ldots,x_N)$ supplies supervision, while $\mathbf z=(z_1,\ldots,z_N)$ denotes a generated path. The parameter vector $\omega=(\varphi,\phi,\psi)$ collects all trainable parameters. 

The objective combines reconstruction accuracy, a transition-density term, agreement between training segments, and a distributional score for full trajectories.

\subsubsection{Reconstruction and Transition Terms}
With the supplied initial state, observations, and controls fixed, $q_\omega(\mathbf z)$ denotes the generated path density. To score a generated path against the reference path by a log density, we use the Gaussian reconstruction model $p_\sigma(x_k\mid z_k)=\N(x_k;z_k,\sigma_{\mathrm{state}}^2I_d)$ with a fixed scale $\sigma_{\mathrm{state}}>0$. Its only role is to make $\mathcal L_{\mathrm{state}}+\mathcal L_{\mathrm{trans}}$ below the expected negative log joint density of the reference and generated paths (Appendix~\ref{app:training-loss}). In practice, it amounts to a squared state error whose weight relative to the entropy term is set by the hyperparameter $\sigma_{\mathrm{state}}$. The reconstruction and transition losses are
\begin{align}
 \mathcal L_{\mathrm{state}}(\omega;\mathbf x)
 &=\frac{1}{2\sigma_{\mathrm{state}}^2}
   \E_{\mathbf z\sim q_\omega}
   \left[\sum_{k=1}^N\norm{x_k-z_k}^2\right]
   +\frac{Nd}{2}\log(2\pi\sigma_{\mathrm{state}}^2),\label{eq:state-training-loss}\\
 \mathcal L_{\mathrm{trans}}(\omega)
 &=\E_{\mathbf z\sim q_\omega}[-\log q_\omega(\mathbf z)]
   =H(q_\omega),\label{eq:transition-training-loss}
\end{align}
where $H(q_\omega)$ is the differential entropy of the generated path. The reconstruction scale is distinct from observation noise and learned transition uncertainty.

Both expectations are taken over generated paths, which are not available in closed form. Evaluating $\mathcal L_{\mathrm{state}}$ therefore requires integrating Equation~\eqref{eq:normflowsde-overview} forward from the supplied initial state, one step per physical interval, with a fresh Gaussian input at every step. Equation~\eqref{eq:normflowsde-overview} is itself an Euler--Maruyama-type update \citep{kloeden1992numerical}, and this is the sampler we use on all four datasets. Higher-order stochastic solvers can also be applied after adapting the stochastic term, such as the Heun \citep{rumelin1982numerical,kloeden1992numerical}, reversible Heun \citep{kidger2021efficient}, and midpoint \citep{milstein2004stochastic} solvers.

For a differentiable, invertible flow with nonsingular forward Jacobian $J_{\phi,k}(\epsilon)=\partial g_\phi(\epsilon;z_k,t_k,u_k,c_k)/\partial\epsilon$, change of variables in Equation~\eqref{eq:normflowsde-overview} gives
\begin{equation}\label{eq:sampled-transition-density}
 \log q_\omega(z_{k+1}\mid z_k,t_k,u_k,c_k)
 =\log\N(\epsilon_k;0,I_d)
  -\log|\det J_{\phi,k}(\epsilon_k)|-\frac d2\log\Delta t.
\end{equation}
The Gaussian log density has a parameter-independent expectation because $\epsilon_k\sim\N(0,I_d)$. Thus only the expected sum of forward log determinants in $\mathcal L_{\mathrm{trans}}$ affects optimization.

Minimizing this entropy term favors concentrated transitions. The conditional flow is built from the affine coupling layers of Equation~\eqref{eq:affine-coupling}, conditioned on $(z_k,t_k,u_k,c_k)$ through the feature $\chi_k$ of Equation~\eqref{eq:shared-feature}. To keep the transitions nondegenerate, we bound their log scales to $(-\alpha,\alpha)$ by setting $\kappa_\ell=\alpha\tanh(\cdot)$, where the bound $\alpha>0$ is a fixed hyperparameter (Table~\ref{tab:nfsde-arch}). For a flow with $L_{\mathrm f}$ coupling layers, where layer $\ell$ rescales $d_\ell$ coordinates, this bounds the Jacobian determinant on both sides:
\begin{equation}\label{eq:training-determinant-bound}
 e^{-B}\leq|\det J_{\phi,k}(\epsilon)|\leq e^B,
 \qquad B=\alpha\sum_{\ell=1}^{L_{\mathrm f}} d_\ell,
 \qquad \epsilon\in\R^d.
\end{equation}
The lower side bounds the transition entropy below, so the entropy term cannot collapse the stochastic increments. As a side effect, it also imposes a noise floor that is set by $\alpha$ and $L_{\mathrm f}$. A learnable per-coordinate factor is therefore required. We denote the bounded coupling flow described so far by $\bar g_\phi$ and define the noise map of Equation~\eqref{eq:normflowsde-overview} as its product with a learned scale,
\begin{equation}\label{eq:learned-noise-scale}
 g_\phi(\epsilon;z_k,t_k,u_k,c_k)
 =\mathrm{diag}(s_\phi)\,\bar g_\phi(\epsilon;z_k,t_k,u_k,c_k).
\end{equation}
The scale $s_\phi\in\R^d_{>0}$ consists of $d$ free parameters, one per state coordinate. It does not depend on the state, the time, the control, or the observations, so the dependence of the noise on these inputs is carried by the coupling layers alone. Its logarithm is bounded within a much wider range than the coupling log scales, and the scale is initialized to one (Appendix~\ref{app:training-loss}).

The sum $\mathcal L_{\mathrm{state}}+\mathcal L_{\mathrm{trans}}$ forms a composite reconstruction-and-entropy objective. Appendix~\ref{app:training-loss} gives the derivation together with the determinant and entropy bounds.

\subsubsection{Multiple Shooting and Boundary Consistency}
To shorten the trajectories through which gradients must be propagated, we choose an integer segment length $L_S\geq2$, counted in state points. This count includes the supplied initial state, so each full segment advances $L_S-1$ time steps and ends at local index $k=L_S-1$. We divide the $N$ transitions into $S$ segments, including a shorter final segment when needed:
\begin{equation}\label{eq:segment-partition}
 S=\left\lceil\frac{N}{L_S-1}\right\rceil,\qquad
 \sum_{s=1}^S(L^{(s)}-1)=N,\qquad
 L^{(s)}=1+\min\{L_S-1,\,N-(s-1)(L_S-1)\},\qquad s=1,\ldots,S.
\end{equation}
Here $s$ labels the segment and $L^{(s)}$ is its actual number of state points. Local index $k$ in segment $s$ corresponds to global time index $(s-1)(L_S-1)+k$.

The supplied initial state of each segment initializes its training rollout. We generate the segments independently in parallel, with the same parameters $\omega$. The boundary penalty compares the predicted endpoint of each nonfinal segment at $k=L_S-1$ with the next supplied initial state at $k=0$:
\begin{equation}\label{eq:continuity-training-loss}
 \mathcal L_{\mathrm{cont}}(\omega)
 =\E
 \left[\sum_{s=1}^{S-1}\norm{z_{L_S-1}^{(s)}-z_0^{(s+1)}}^2\right].
\end{equation}
The expectation is estimated by averaging the summed squared boundary discrepancies over sampled segment rollouts.

\subsubsection{Full-Horizon Distributional Supervision}\label{sec:training-rollout}
To account for errors that accumulate over the complete horizon, we generate full trajectories from the supplied initial state $z_0$, without restarting at segment boundaries. The vectors $\mathbf z$ and $\mathbf z'$ denote two such trajectories generated using independent Gaussian inputs, with the same observations, controls, and initial state. We compare them with the reference trajectory $\mathbf x$ using the energy score \citep{gneiting2007proper}:
\begin{equation}\label{eq:flow_sde_rollout_loss}
 \mathcal L_{\mathrm{rollout}}(\omega;\mathbf x)
 =\E\,\left[\norm{\mathbf z-\mathbf x}\right]
 -\frac12\E\,\left[\norm{\mathbf z-\mathbf z'}\right].
\end{equation}
Each norm is the unsquared Euclidean distance over all time steps and state coordinates. The expectations average over the generated trajectories, holding the supplied data fixed. The first term encourages agreement with the reference trajectory. The second rewards separation between generated trajectories and counteracts their collapse to a single prediction.

During training, we generate $M\geq2$ independent full trajectories $\mathbf z^{(1)},\ldots,\mathbf z^{(M)}$ and compute
\begin{equation}\label{eq:energy-score-estimator}
 \widehat{\mathcal L}_{\mathrm{rollout}}(\omega;\mathbf x)
 =\frac1M\sum_{i=1}^M\norm{\mathbf z^{(i)}-\mathbf x}
 -\frac{1}{2M(M-1)}\sum_{i\ne j}\norm{\mathbf z^{(i)}-\mathbf z^{(j)}}.
\end{equation}
This estimator averages distances to the reference and subtracts half the average distance between distinct generated trajectories. The second sum includes both orders of each pair and excludes self-pairs, giving an unbiased estimate of Equation~\eqref{eq:flow_sde_rollout_loss}.

For fixed observations, controls, and initial state, the expected energy score over reference trajectories is uniquely minimized by matching their true conditional distribution, provided the reference and generated trajectories have finite expected norms \citep{gneiting2007proper,szekely2013energy}. The energy score helps maintain diversity among generated samples while encouraging agreement with the reference trajectory.

\subsubsection{Combined Objective and Optimization}
The combined training objective is
\begin{equation}\label{eq:normflowsde_loss}
 \E_{\mathcal D_{\mathrm{train}}}
 [\mathcal L(\omega;\mathbf x)] = \E_{\mathcal D_{\mathrm{train}}}
 [\mathcal L_{\mathrm{state}}(\omega;\mathbf x)
 +\mathcal L_{\mathrm{trans}}(\omega)+\lambda_{\mathrm{cont}}\mathcal L_{\mathrm{cont}}(\omega)
 +\lambda_{\mathrm{rollout}}\mathcal L_{\mathrm{rollout}}(\omega;\mathbf x)].
 \end{equation}
Here $\E_{\mathcal D_{\mathrm{train}}}$ averages over paired reference states, raw observations, controls, and supplied initial states, together with the selected segment partitions and anchors. We evaluate the rollout term every five optimizer updates to reduce the computational cost. 

\providecommand{\tbd}{\textbf{[TBD]}}
\providecommand{\figplaceholder}[1]{\fbox{\parbox[c][0.28\linewidth][c]{0.96\linewidth}{\centering\small Figure placeholder: #1}}}

\section{Empirical evaluation}\label{sec:experiments}
We evaluate NormFlowSDE as a causal amortized Bayesian filter on two simulated systems and two robot datasets against seven learned estimators trained on the same data. Following Section~\ref{sec:related}, the eight methods form two groups of four: direct estimators, which sample the state trajectory conditioned on encoded measurements, and particle filters, which weight their particles with a learned measurement likelihood. One of the particle filters uses NormFlowSDE's own transition and separates the transition model from the direct update (Section~\ref{sec:experiments-methods}). All four datasets are observed through the same heavy-tailed, multimodal measurement noise, so that the comparison also tests how each method copes with a non-Gaussian likelihood. We evaluate the accuracy of the filtering distributions, the calibration of their uncertainty intervals, and the inference cost. All methods have the same information access, use the same data splits and update budgets (with one exception for NormFlowSDE on REASSEMBLE, stated in Section~\ref{sec:experiments-results}), and are scored with the same metrics. Every number in this section is produced from the frozen run directories by the scripts listed in Appendix~\ref{app:reproducibility}.

\subsection{Datasets}\label{sec:experiments-data}
Table~\ref{tab:datasets} summarizes the four datasets. Each sequence consists of a known initial state, a control sequence, and noisy observations of the subsequent states. The reference states are the states before the synthetic observation noise of Equation~\eqref{eq:obsnoise} is added. They supervise training and are used for scoring. However, no method sees them at test time, where only the initial state, the controls, and the noisy observations are available. In real-world applications, these labels come from simulation or from a more accurate reference sensor available during data collection. Every dataset is split by trajectory or episode before windowing, and normalization statistics are computed on the training split only.

\paragraph{Observation noise.}
Each observed coordinate is corrupted independently by the same three-component Gaussian mixture,
\begin{equation}\label{eq:obsnoise}
 V\sim 0.70\,\N\,\big(0,(0.5\sigma)^2\big)+0.15\,\N\,\big(2.5\sigma,(1.5\sigma)^2\big)+0.15\,\N\,\big(-2.5\sigma,(1.5\sigma)^2\big),
\end{equation}
whose scale $\sigma$ is 0.1 for the pendulum (in rad and rad/s), 0.5 for Lorenz--96, and 0.1 in normalized units for the two robot datasets. The mixture has standard deviation $1.65\,\sigma$. We compare a reconstructed noise density $q$ with the mixture density $p$ by the overlap coefficient
\begin{equation}\label{eq:overlap}
 \mathrm{OVL}(p,q)=\int\min\{p(v),q(v)\}\,dv=1-\TV(p,q),
\end{equation}
which equals one for identical densities and zero for densities with disjoint support. The Gaussian with the same mean and variance as the mixture has an overlap of 0.668 with it, which serves as a reference value for a Gaussian noise model.

\begin{table}[t]
\centering\footnotesize
\begin{tabular}{lllll}
\toprule
 & Pendulum & Lorenz--96 & ALOHA insertion & REASSEMBLE \\
\midrule
Source & simulated & simulated & sim.\ demonstrations & real robot \\
State dimension $d$ & 2 & 12 & 14 & 7 \\
Observed coordinates & 2 & 6 & 14 & 7 \\
Control dimension & 1 & -- & 14 & 9 \\
Observation noise & mixture, $\sigma=0.1$ & mixture, $\sigma=0.5$ & mixture, $\sigma=0.1$ & mixture, $\sigma=0.1$ \\
Time step $\Delta t$ & 0.05 & 0.02 & 0.02 & 1/30 \\
Window length (steps) & 60 & 100 & 250 & 250 \\
Train / val.\ / test windows & 224 / 48 / 48 & 1600 / 200 / 100 & 400 / 50 / 50 & 2000 / 300 / 300 \\
Samples / particles $M$ & 128 & 200 & 200 & 500 \\
Metric units & physical & physical & normalized & normalized \\
\bottomrule
\end{tabular}
\caption{Datasets. The observation noise is the mixture of Equation~\eqref{eq:obsnoise} with the stated scale $\sigma$. The label ``sim.\ demonstrations'' refers to human teleoperation recorded in simulation. $M$ is the number of posterior draws scored per window (Appendix~\ref{app:protocol}).}
\label{tab:datasets}
\end{table}

\paragraph{Controlled pendulum.}
The first system is a damped pendulum driven by a scalar torque $u$. It is simulated as the It\^o SDE
\begin{equation}\label{eq:pendulum}
 d\theta=\omega\,dt+\sigma_\theta\,dW_\theta,\qquad
 d\omega=\Big(\frac{u}{m\ell^{2}}-c\,\omega-\frac{g}{\ell}\sin\theta\Big)dt+\sigma_\omega\,dW_\omega,
\end{equation}
with $g=9.81$, $\ell=m=1$, damping $c=0.2$, and process-noise scales $\sigma_\theta=0.01$ and $\sigma_\omega=0.2$. The SDE is integrated by Euler--Maruyama with eight substeps per observation interval $\Delta t=0.05$. The initial state is known to every filter and is drawn as $\theta_0\sim\U(-\pi,\pi)$, $\omega_0\sim\N(0,0.6^{2})$. Both coordinates are observed. 

\paragraph{Lorenz--96.}
The second system is the twelve-dimensional Lorenz--96 model,
\begin{equation}\label{eq:lorenz96}
 \frac{dx_i}{dt}=(x_{i+1}-x_{i-2})x_{i-1}-x_i+F,
 \qquad i=0,\ldots,11,\qquad F=8,
\end{equation}
with cyclic indices. It is integrated with a fourth-order Runge--Kutta scheme at $\Delta t=0.02$. Each window is an independent trajectory. Its initial state is drawn as $x(0)\sim\N(0,\,0.01^{2}I)$ and integrated for a spin-up of 500 steps so that it lies on the attractor. The state at the end of the spin-up is the known initial state $x_0$ of the window. Only every second coordinate is observed, so six coordinates must be inferred from the dynamics alone. 

\paragraph{ALOHA insertion.}
The third dataset consists of bimanual peg-insertion demonstrations from the ALOHA simulation benchmark, recorded by human teleoperation at 50\,Hz \citep{zhao2023aloha}. Each frame stores two different 14-dimensional joint vectors, which we keep apart. The \emph{commanded} joint positions are the leader-arm targets sent to the position controllers of the follower arms, and they serve as the control input. The \emph{measured} joint positions of the follower arms define the state. The two differ by the tracking error of the controllers, which is largest during contact. The recorded states are deterministic given the commands, so we perturb them by a slowly varying latent disturbance before normalization. This disturbance is a four-dimensional AR(1) velocity offset whose magnitude grows with the tracking error and is therefore largest during contact. The exact recipe is given in Appendix~\ref{app:reproducibility}. The perturbed trajectory is the reference state, and it is observed through the noise of Equation~\eqref{eq:obsnoise} with $\sigma=0.1$ in normalized units. 

\paragraph{REASSEMBLE.}
The fourth dataset consists of contact-rich assembly and disassembly episodes recorded on a real seven-degree-of-freedom robot arm at 30\,Hz \citep{sliwowski2025reassemble}. The state consists of the seven joint positions, all of which are observed. The nine-dimensional robot commands serve as controls. Unlike the other datasets, the dynamics are those of a physical system interacting with its environment, and no model of them is available.

\subsection{Compared Methods}\label{sec:experiments-methods}
All eight methods are causal: they start from the known initial state, and no estimate uses a measurement taken after its own time. The particle filters weight the prediction of $x_{k+1}$ with $y_{k+1}$. In contrast, the direct estimators produce $x_{k+1}$ from $x_k$ and the window feature $c_k$ of Equation~\eqref{eq:sequence-conditioning}, which ends at $y_k$. The \emph{direct estimators} are NormFlowSDE, latent NSF, latent SDE, and SigKer SDE. The three baselines use the encoder of NormFlowSDE (the same window of ten measurements and the same width) and have the parameter count of NormFlowSDE. Because our data pair states with measurements, their latent state is identified with the physical state and supervised by the reference states (Appendix~\ref{app:direct-baselines}). The \emph{particle filters} (Section~\ref{sec:related-pf}) obtain their uncertainty through Bayesian weighting with learned transition and measurement models. NSF-PF, FlowSDE-PF, and NormFlowSDE-PF are bootstrap filters whose measurement flows have the same architecture and training, each fitted jointly with its own transition, so that they differ only in the transition. NF-DPF additionally samples its particles from a measurement-conditioned proposal. NormFlowSDE-PF uses the transition of Equation~\eqref{eq:normflowsde-overview}, conditioned on the control and the time but not on observations, with the particle counts and the training and selection recipe of NSF-PF (Appendix~\ref{app:normflowsde-pf}). It separates the transition model of NormFlowSDE from the way it is used. For the two simulated systems, pendulum and Lorenz--96, we also report model-based reference filters that use the true dynamics and the true noise scale: an extended Kalman filter (EKF) and an unscented Kalman filter (UKF) with the moment-matched Gaussian likelihood, and a bootstrap particle filter with the exact mixture likelihood. These references indicate the accuracy attainable with perfect models and are not competitors.

\subsection{Benchmark Results}\label{sec:experiments-results}
Tables~\ref{tab:results-sim} and~\ref{tab:results-robot} compare the eight methods on the simulated systems and the robot datasets. Each configuration is trained with three independent seeds, and we report the mean and standard deviation over seeds. Accuracy is measured by the RMSE of the posterior mean, averaged over test windows. Distributional accuracy is measured by the continuous ranked probability score (CRPS) and the energy score \citep{gneiting2007proper}, averaged over coordinates. Calibration is measured by the empirical coverage of central 90\% marginal intervals, pooled over windows, steps and coordinates, which is read together with their mean width. Inference cost is the median wall-clock time to filter one test window on a single GPU, measured for every method on the same device. 

\begin{table}[htbp]
\centering\footnotesize
\resizebox{\linewidth}{!}{%
\begin{tabular}{lllccccc}
\toprule
Dataset & Route & Method & RMSE $\downarrow$ & CRPS $\downarrow$ & Cov.\ 90\% & Width & Latency (ms) $\downarrow$ \\
\midrule
Pendulum & direct & NormFlowSDE & 0.0736 $\pm$ 0.0029 & 0.0387 $\pm$ 0.0012 & 0.920 $\pm$ 0.002 & 0.2321 $\pm$ 0.0061 & 2.0 \\
 &  & Latent NSF & 0.1534 $\pm$ 0.0325 & 0.0666 $\pm$ 0.0100 & 0.889 $\pm$ 0.004 & 0.3224 $\pm$ 0.0242 & \textbf{1.7} \\
 &  & Latent SDE & 0.1063 $\pm$ 0.0024 & 0.0531 $\pm$ 0.0021 & 0.933 $\pm$ 0.012 & 0.3388 $\pm$ 0.0069 & 2.0 \\
 &  & SigKer SDE & 0.1815 $\pm$ 0.0173 & 0.0993 $\pm$ 0.0107 & 0.174 $\pm$ 0.080 & 0.1068 $\pm$ 0.0321 & 2.0 \\
\cmidrule(l){2-8}
 & PF & NF-DPF & 0.0792 $\pm$ 0.0053 & 0.0392 $\pm$ 0.0021 & 0.913 $\pm$ 0.004 & 0.2500 $\pm$ 0.0192 & 8.5 \\
 &  & NSF-PF & \textbf{0.0650 $\pm$ 0.0004} & \textbf{0.0333 $\pm$ 0.0001} & 0.858 $\pm$ 0.006 & 0.1767 $\pm$ 0.0032 & 8.6 \\
 &  & FlowSDE-PF & 0.1190 $\pm$ 0.0017 & 0.0625 $\pm$ 0.0006 & \textbf{0.906 $\pm$ 0.002} & 0.3554 $\pm$ 0.0227 & 10.9 \\
 &  & NormFlowSDE-PF & 0.0959 $\pm$ 0.0058 & 0.0451 $\pm$ 0.0023 & 0.875 $\pm$ 0.003 & 0.2158 $\pm$ 0.0059 & 8.2 \\
\cmidrule(l){2-8}
 & ref. & EKF / UKF & 0.0578 $\pm$ 0.0000 & 0.0296 $\pm$ 0.0000 & 0.870 $\pm$ 0.001 & 0.1650 $\pm$ 0.0002 & 11.5 / 11.7 \\
 &  & bootstrap PF & 0.0406 $\pm$ 0.0003 & 0.0192 $\pm$ 0.0001 & 0.895 $\pm$ 0.002 & 0.1129 $\pm$ 0.0000 & 10.4 \\
\midrule
Lorenz--96 & direct & NormFlowSDE & 0.2725 $\pm$ 0.0019 & 0.1464 $\pm$ 0.0010 & 0.910 $\pm$ 0.004 & 0.8767 $\pm$ 0.0059 & 4.9 \\
 &  & Latent NSF & 0.8230 $\pm$ 0.0147 & 0.4021 $\pm$ 0.0040 & 0.875 $\pm$ 0.003 & 2.1569 $\pm$ 0.0341 & \textbf{2.9} \\
 &  & Latent SDE & 0.3304 $\pm$ 0.0061 & 0.1836 $\pm$ 0.0037 & 0.800 $\pm$ 0.006 & 0.7675 $\pm$ 0.0054 & 4.2 \\
 &  & SigKer SDE & 0.4536 $\pm$ 0.0071 & 0.2645 $\pm$ 0.0069 & 0.542 $\pm$ 0.034 & 0.6036 $\pm$ 0.0461 & 4.4 \\
\cmidrule(l){2-8}
 & PF & NF-DPF & 0.2724 $\pm$ 0.0073 & 0.1476 $\pm$ 0.0038 & 0.883 $\pm$ 0.012 & 0.7798 $\pm$ 0.0434 & 28.2 \\
 &  & NSF-PF & \textbf{0.1556 $\pm$ 0.0046} & \textbf{0.0783 $\pm$ 0.0021} & 0.921 $\pm$ 0.005 & 0.4731 $\pm$ 0.0009 & 20.8 \\
 &  & FlowSDE-PF & 0.3017 $\pm$ 0.0038 & 0.1634 $\pm$ 0.0026 & \textbf{0.899 $\pm$ 0.002} & 0.9349 $\pm$ 0.0012 & 30.1 \\
 &  & NormFlowSDE-PF & 0.1974 $\pm$ 0.0071 & 0.1011 $\pm$ 0.0040 & 0.876 $\pm$ 0.003 & 0.4759 $\pm$ 0.0025 & 20.6 \\
\cmidrule(l){2-8}
 & ref. & EKF / UKF & 0.0229 $\pm$ 0.0000 & 0.0093 $\pm$ 0.0000 & 0.997 $\pm$ 0.000 & 0.0906 $\pm$ 0.0001 & 12.0 / 18.1 \\
 &  & bootstrap PF & 0.0188 $\pm$ 0.0001 & 0.0077 $\pm$ 0.0000 & 0.990 $\pm$ 0.000 & 0.0674 $\pm$ 0.0000 & 15.5 \\
\bottomrule
\end{tabular}

}
\caption{Test results on the simulated systems for the four direct estimators and the four particle filters (PF). Entries are the mean $\pm$ standard deviation over three training seeds. Each method returns $M$ posterior draws per window as in Table~\ref{tab:datasets}, and the particle filters run with the particle count of the dataset protocol. Bold marks the best learned method per dataset, which for coverage is the one closest to the nominal level of 0.90. RMSE and width are in the metric units of Table~\ref{tab:datasets}. Latency is the median time in milliseconds to filter one test window, measured for seed 0 on one NVIDIA L40S with compilation excluded (Appendix~\ref{app:protocol}), and bold marks the fastest learned method. The rows labeled ``ref.'' are model-based reference filters with the true dynamics, which are not competitors.}
\label{tab:results-sim}
\end{table}

\paragraph{Simulated systems.}
On the pendulum and Lorenz--96, the most accurate learned method is a particle filter. NSF-PF reaches the lowest RMSE and CRPS on both systems. On the pendulum, NormFlowSDE is the second most accurate method (0.074 against 0.065) and more accurate than the other three particle filters. On Lorenz--96, NSF-PF (0.156) and NormFlowSDE-PF (0.197) are clearly ahead, and NormFlowSDE (0.273) is as accurate as NF-DPF (0.272) and more accurate than FlowSDE-PF. NormFlowSDE is the most accurate direct estimator on both systems. Both systems have smooth, low-dimensional dynamics that a one-step transition model can learn accurately from clean state pairs. Once these models are accurate, Bayesian weighting is close to optimal. The reference filters put the numbers in perspective. On the pendulum, the exact-likelihood particle filter with the true model reaches 0.041. However, the Gaussian filters, including EKF and UKF, which are with the true dynamics but the moment-matched likelihood,  reach only 0.058. About 70\% of the gap between the best learned filter and the exact-likelihood reference is therefore caused by the noise model alone. On Lorenz--96 the reference particle filter reaches 0.019, an order of magnitude below every learned method. This is due to the exact chaotic dynamics whereas the other learned state estimators have only 6 out of 12 observable states. 

Judged by coverage and interval width together, NormFlowSDE is the only direct estimator whose 90\% intervals are both calibrated and informative on the two systems: its coverage is within 0.02 of the nominal level (0.920 and 0.910). FlowSDE-PF is closer to nominal (0.906 and 0.899), but its intervals are 53\% and 7\% wider than those of NormFlowSDE and its RMSE is higher, so it is calibrated but not sharp. The same holds for latent NSF, whose near-nominal coverage (0.889 and 0.875) rests on intervals 1.4 and 2.5 times as wide as those of NormFlowSDE. Conversely, narrow intervals do not guarantee coverage: NSF-PF, the most accurate method, under-covers on the pendulum (0.858), and NormFlowSDE-PF under-covers on both systems (0.875 and 0.876).

\begin{table}[htbp]
\centering\footnotesize
\resizebox{\linewidth}{!}{%
\begin{tabular}{lllccccc}
\toprule
Dataset & Route & Method & RMSE $\downarrow$ & CRPS $\downarrow$ & Cov.\ 90\% & Width & Latency (ms) $\downarrow$ \\
\midrule
ALOHA & direct & NormFlowSDE & \textbf{0.0523 $\pm$ 0.0001} & \textbf{0.0291 $\pm$ 0.0001} & \textbf{0.911 $\pm$ 0.008} & 0.1787 $\pm$ 0.0066 & 9.7 \\
 &  & Latent NSF & 0.0559 $\pm$ 0.0018 & 0.0308 $\pm$ 0.0011 & 0.878 $\pm$ 0.021 & 0.1709 $\pm$ 0.0133 & \textbf{5.8} \\
 &  & Latent SDE & 0.0597 $\pm$ 0.0006 & 0.0332 $\pm$ 0.0004 & 0.884 $\pm$ 0.022 & 0.1856 $\pm$ 0.0133 & 8.7 \\
 &  & SigKer SDE & 0.0648 $\pm$ 0.0015 & 0.0426 $\pm$ 0.0039 & 0.250 $\pm$ 0.158 & 0.0372 $\pm$ 0.0249 & 9.1 \\
\cmidrule(l){2-8}
 & PF & NF-DPF & 0.0969 $\pm$ 0.0044 & 0.0565 $\pm$ 0.0026 & 0.751 $\pm$ 0.009 & 0.2198 $\pm$ 0.0046 & 50.2 \\
 &  & NSF-PF & 0.0828 $\pm$ 0.0006 & 0.0484 $\pm$ 0.0003 & 0.830 $\pm$ 0.001 & 0.2321 $\pm$ 0.0004 & 38.2 \\
 &  & FlowSDE-PF & 0.0708 $\pm$ 0.0008 & 0.0408 $\pm$ 0.0005 & 0.868 $\pm$ 0.003 & 0.2178 $\pm$ 0.0022 & 48.4 \\
 &  & NormFlowSDE-PF & 0.0778 $\pm$ 0.0002 & 0.0453 $\pm$ 0.0002 & 0.854 $\pm$ 0.001 & 0.2319 $\pm$ 0.0007 & 37.8 \\
\midrule
REASSEMBLE & direct & NormFlowSDE & \textbf{0.0319 $\pm$ 0.0008} & \textbf{0.0176 $\pm$ 0.0004} & 0.912 $\pm$ 0.001 & 0.1053 $\pm$ 0.0018 & 10.3 \\
 &  & Latent NSF & 0.0335 $\pm$ 0.0005 & 0.0180 $\pm$ 0.0003 & \textbf{0.890 $\pm$ 0.001} & 0.0992 $\pm$ 0.0014 & \textbf{6.4} \\
 &  & Latent SDE & 0.0365 $\pm$ 0.0004 & 0.0199 $\pm$ 0.0002 & 0.918 $\pm$ 0.002 & 0.1227 $\pm$ 0.0006 & 9.8 \\
 &  & SigKer SDE & 0.0494 $\pm$ 0.0073 & 0.0307 $\pm$ 0.0022 & 0.382 $\pm$ 0.202 & 0.0519 $\pm$ 0.0373 & 9.8 \\
\cmidrule(l){2-8}
 & PF & NF-DPF & 0.0598 $\pm$ 0.0007 & 0.0331 $\pm$ 0.0004 & 0.873 $\pm$ 0.008 & 0.1791 $\pm$ 0.0027 & 48.8 \\
 &  & NSF-PF & 0.0612 $\pm$ 0.0002 & 0.0343 $\pm$ 0.0001 & 0.917 $\pm$ 0.000 & 0.2138 $\pm$ 0.0001 & 44.0 \\
 &  & FlowSDE-PF & 0.0526 $\pm$ 0.0014 & 0.0297 $\pm$ 0.0009 & 0.952 $\pm$ 0.002 & 0.2114 $\pm$ 0.0052 & 54.8 \\
 &  & NormFlowSDE-PF & 0.0618 $\pm$ 0.0006 & 0.0348 $\pm$ 0.0004 & 0.919 $\pm$ 0.002 & 0.2188 $\pm$ 0.0045 & 41.8 \\
\bottomrule
\end{tabular}

}
\caption{Test results on the robot datasets, in the format of Table~\ref{tab:results-sim}.}
\label{tab:results-robot}
\end{table}

\paragraph{Robot datasets.}
On ALOHA and REASSEMBLE the direct route is the more accurate one. On both datasets every direct estimator is more accurate than the corresponding particle filter. The RMSE of NormFlowSDE is 26\% lower than that of the best particle filter, FlowSDE-PF, on ALOHA and 39\% lower on REASSEMBLE. Here the transition dynamics are high-dimensional, and for REASSEMBLE they are those of a physical, contact-rich system. The particle filters rely on a learned one-step transition model and a learned measurement density, and errors in either directly affect their weights. NF-DPF in particular loses accuracy and coverage on ALOHA. The direct estimators instead learn the observation-conditioned update and use every observed coordinate without an explicit likelihood. Within the direct group, NormFlowSDE attains the lowest RMSE and CRPS on both datasets, with an RMSE 6\% and 5\% below that of latent NSF. NormFlowSDE, latent NSF, and the latent SDE all cover within 0.025 of the nominal level on both datasets (0.88 to 0.92), which no particle filter achieves. NormFlowSDE is the closest of the three on ALOHA (0.91), and latent NSF is slightly closer on REASSEMBLE (0.89 against 0.91). Among these three, latent NSF has the narrowest intervals on both datasets. The signature-kernel SDE is strongly under-dispersed.

\begin{figure}[htbp]
\centering
\includegraphics[width=\linewidth]{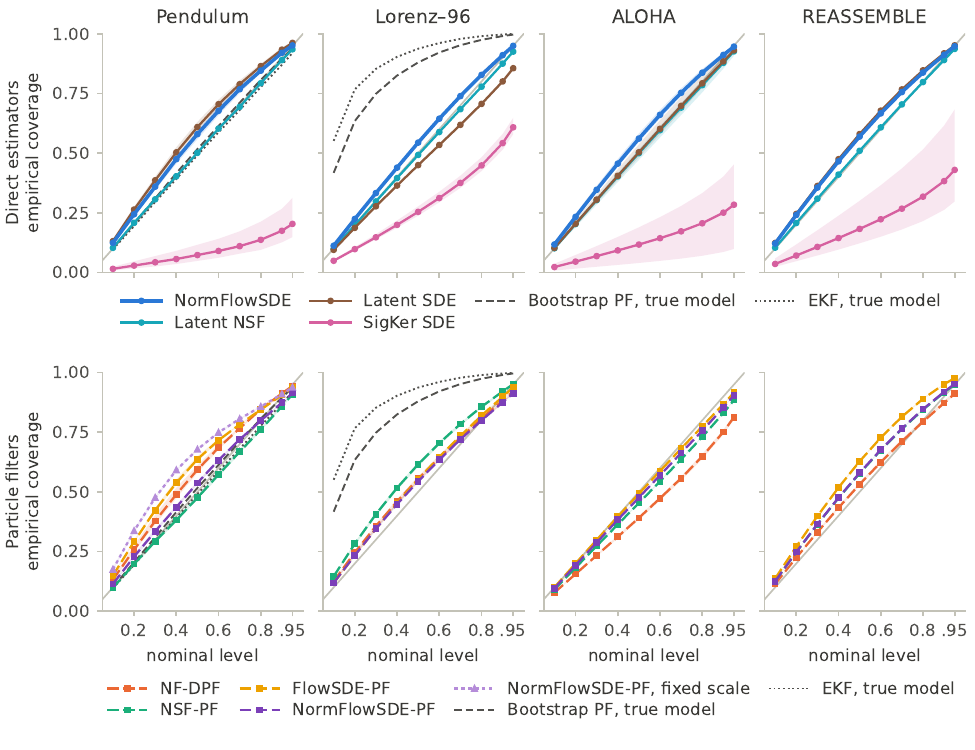}
\caption{Calibration of central marginal intervals at nominal levels from 10\% to 95\%, pooled over test windows, steps, and coordinates. The curves show the mean over three seeds, and the shaded bands show the range over seeds. The top row shows the direct estimators. The bottom row shows the particle filters, including the pendulum ablation of NormFlowSDE-PF without the learned scale. On REASSEMBLE, the curves of NSF-PF and NormFlowSDE-PF coincide. The diagonal indicates perfect calibration. The gray curves are the reference filters with the true model on the simulated systems.}
\label{fig:calibration}
\end{figure}

\paragraph{Calibration.}
Figure~\ref{fig:calibration} shows the empirical coverage of central intervals at nominal levels from 10\% to 95\% for all eight learned methods, with the direct estimators in the top row and the particle filters in the bottom row. The intervals of NormFlowSDE are close to nominal at the high levels on all four datasets (within 0.02 at the 90\% level and within 0.005 at the 95\% level). At the central levels they are slightly too wide: its coverage at the 50\% level is 0.54 to 0.58. Among the direct baselines, latent NSF lies closest to the diagonal on every dataset (within 0.025 at every level). The latent SDE is calibrated on ALOHA, over-dispersed on the pendulum and REASSEMBLE, and under-dispersed on Lorenz--96. The signature-kernel SDE is strongly under-dispersed on every dataset. 

The particle filters are miscalibrated in both directions depending on the dataset. NF-DPF is over-dispersed at the central levels on the pendulum and strongly under-dispersed on ALOHA. FlowSDE-PF is nominal at the 90\% level on both simulated systems but over-dispersed at the central levels on the pendulum (0.64 at the 50\% level), it is over-dispersed at every level on REASSEMBLE, and it is the best calibrated particle filter on ALOHA (within 0.035 at every level). NSF-PF is under-dispersed on the pendulum and ALOHA and over-dispersed at the central levels on Lorenz--96 and REASSEMBLE. NormFlowSDE-PF is under-dispersed at the high levels on the pendulum and ALOHA and over-dispersed on REASSEMBLE. Among the methods whose 90\% coverage lies within 0.025 of the nominal level on the robot datasets, latent NSF has the narrowest intervals, closely followed by NormFlowSDE. The signature-kernel SDE has even narrower intervals on both datasets, but they come with under-coverage. The reference filters are over-dispersed on Lorenz--96, so even the true model does not yield calibrated intervals on the chaotic system.

\begin{figure}[t]
\centering
\includegraphics[width=\linewidth]{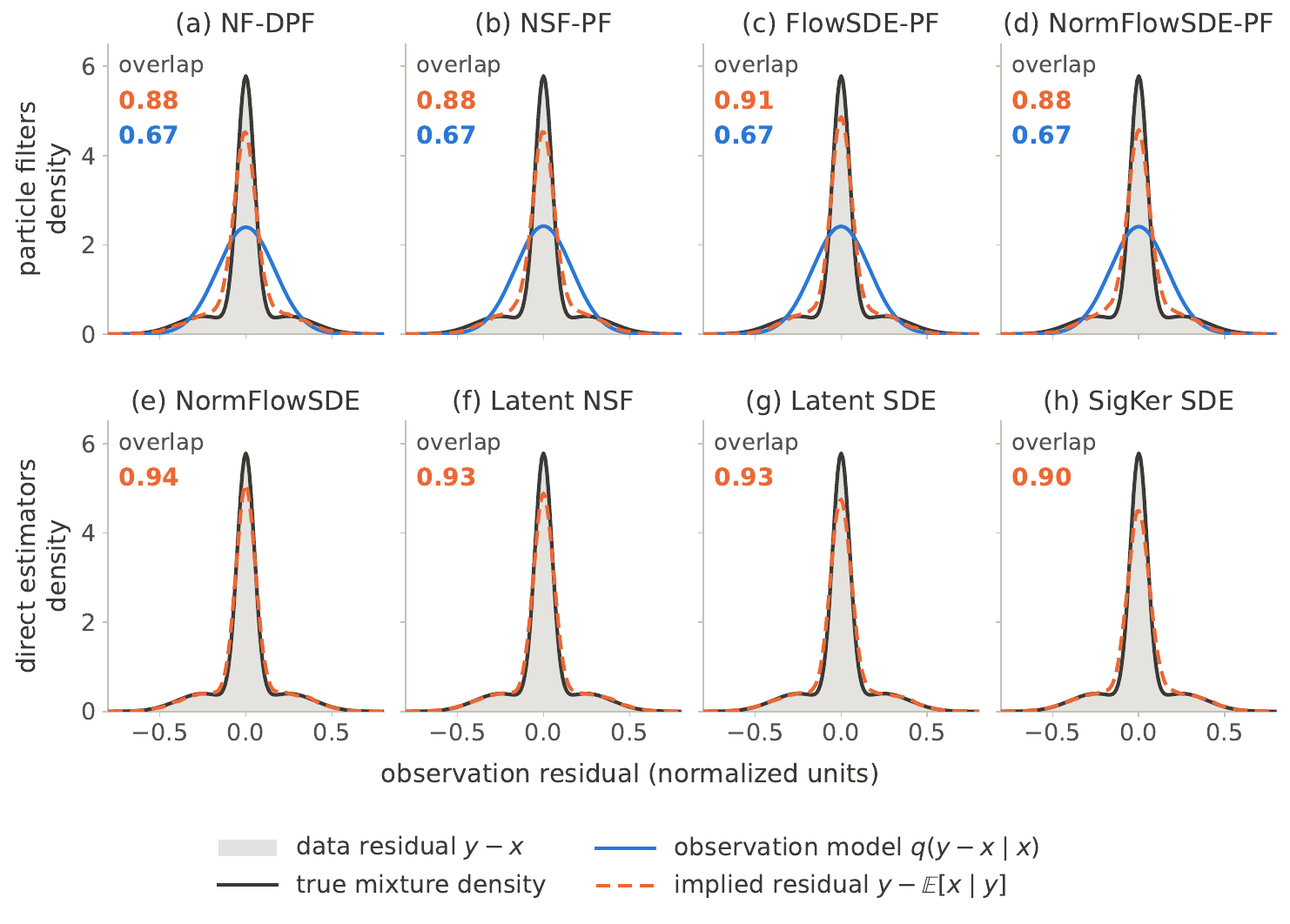}
\caption{Observation noise on REASSEMBLE as represented by the four particle filters (top, a--d) and the four direct estimators (bottom, e--h), shown for seed 0 in normalized units. Gray: histogram of the data residuals $y-x$; black: the mixture of Equation~\eqref{eq:obsnoise}; orange, dashed: the implied residuals $y-\E[x\mid y]$ of each method's posterior mean on all test windows; blue (particle filters only): the learned observation model $q(y\mid x)$ as a density of the residual $y-x$, averaged over test states, with the 10\% to 90\% range over states, which is barely wider than the line. The numbers in each panel are overlap coefficients of Equation~\eqref{eq:overlap} with the mixture, where one means identical densities: in orange for a kernel density estimate of the implied residuals, and in blue for the observation model.}
\label{fig:noise-recovery}
\end{figure}

\paragraph{Shape of the observation noise.}
Figure~\ref{fig:noise-recovery} shows how well each method recovers the mixture noise of Equation~\eqref{eq:obsnoise} on REASSEMBLE. For every method, it compares the distribution of the implied residuals $y-\E[x\mid y]$, that is, the measurements minus the posterior mean, with the true mixture. The implied residuals of the direct estimators keep the spread of the data (standard deviation 0.163 to 0.172 against 0.165) and follow the shape of the mixture. NormFlowSDE has the largest overlap to the ground truth. In contrast, the implied residuals of the particle filters overlap less with the mixture (0.88 to 0.91) and are 6\% to 11\% narrower than the data, because their posterior means partly follow the noisy measurements. The learned measurement flows $q(y\mid x)$ of the particle filters, drawn in blue in panels (a) to (d), explain this difference. All four coincide with the moment-matched Gaussian, and their overlap with the mixture is 0.67, which is the analytic value for that Gaussian. Maximum-likelihood training thus matches the variance of the noise but not its sharp central component or its side modes. The particles are therefore weighted with a Gaussian likelihood, which is the same deficit that limits the EKF and UKF on the pendulum.

\paragraph{Latency.}
A direct estimator requires one network evaluation per time step, without weighting or resampling. A particle filter additionally evaluates the measurement density for every particle at every step and resamples when the effective sample size drops. The latency columns of Tables~\ref{tab:results-sim} and~\ref{tab:results-robot} report the median time to filter one test window in the configuration of each row. All eight learned methods and the reference filters were timed on the same GPU. Every direct estimator is faster than every particle filter on every dataset. NormFlowSDE is faster than the fastest particle filter by a factor of 3.9 to 4.2. The cost advantage is a property of the direct route rather than of NormFlowSDE alone. The direct baselines also need one encoder and update evaluation per step and no likelihood evaluation, and their cost is similar. Latent NSF is the fastest method on every dataset. It is faster than NormFlowSDE by a factor of 1.2 on the pendulum and 1.6 to 1.7 on the other datasets (1.7 to 6.4\,ms against 2.0 to 10.3\,ms), because Latent NSF samples from a Gaussian posterior whereas NormFlowSDE update passes through a conditional coupling flow during inference. NormFlowSDE is within 0.02\,ms of the latent SDE and the signature-kernel SDE on the pendulum and 0.5 to 1.1\,ms slower than them on the other three datasets.

\section{Conclusion}\label{sec:conclusion}
NormFlowSDE replaces the likelihood and importance weights of a particle filter by an observation-conditioned normalizing-flow update that is trained end to end on complete filtering rollouts. On four datasets with heavy-tailed, multimodal measurement noise, against three direct baselines and four learned particle filters, it is the most accurate direct estimator on every dataset and the most accurate method on the two robot datasets. Its 90\% coverage lies within 0.02 of the nominal level on all four datasets, which no other method achieves, and it is 3.9 to 4.2 times faster than the fastest particle filter.

\paragraph{Direct estimation or particle filtering.}
The eight methods differ first in whether the observation enters the estimate through a likelihood and importance weights or through a learned observation-conditioned update, and the benchmark separates two regimes. When the dynamics are simple enough to learn from clean state pairs, weighting is more accurate: on the pendulum and Lorenz--96, the RMSE of the best learned particle filter, NSF-PF, is 12\% and 43\% lower than that of NormFlowSDE. The model-based filters show that the remaining gap to perfect inference is small on the pendulum and dominated by chaotic error amplification on Lorenz--96. When the dynamics are complex, contact-rich, and not governed by a closed-form model, as on ALOHA and REASSEMBLE, the direct route is more accurate: every direct estimator has a lower RMSE than every particle filter, and that of NormFlowSDE is 26\% and 39\% lower than that of the best particle filter.

NormFlowSDE and its corresponding particle filter show where the differences between direct estimation and particle filtering come from. On the robot datasets NormFlowSDE-PF is as accurate as NSF-PF but far behind NormFlowSDE. The advantage of NormFlowSDE is therefore due to the observation-conditioned update, which bypasses the learned transition and observation models. On Lorenz--96 the relation is reversed: the NormFlowSDE transition has a 28\% lower RMSE inside the particle filter (0.197 against 0.273). Importance weighting is thus the better use of a transition that can be learned accurately from clean state pairs when half of the state is unobserved. The contrast is even stronger for NSF: as a direct estimator, latent NSF is the least accurate method on Lorenz--96 (RMSE 0.823), whereas the NSF transition inside a particle filter, NSF-PF, is the most accurate (0.156).

\paragraph{Limitations.}
Partially observed chaotic dynamics remain difficult. On Lorenz--96, the six unobserved coordinates can be inferred only through the dynamics, and without a transition model and importance weights NormFlowSDE reaches an RMSE of 0.27 against 0.019 for the reference particle filter. Training also requires careful tuning. The objective of Equation~\eqref{eq:normflowsde_loss} balances four terms, and coverage is sensitive to the rollout weight $\lambda_{\mathrm{rollout}}$, the state scale $\sigma_{\mathrm{state}}$, the initial noise scale, and the learning rate and length of the second training stage. These were selected on validation trajectories separately for each dataset, and a new system requires the same search.

NormFlowSDE is trained with state supervision, as are the measurement models of all learned particle filters. On a physical system, as in our robot experiments, the states are recorded with the inputs by a reference sensor used only during data collection, such as joint encoders, and the observations come from a cheaper sensor or from added noise that models it. The estimator then targets the reference measurement rather than the exact state: the reference noise is absorbed into the learned process noise and bounds the attainable accuracy, an effect that is negligible only when the reference noise is small relative to the process noise. Synthetic observation noise must also match the deployed sensor, and, like any amortized estimator, NormFlowSDE is reliable only within the conditions covered by its training data.

Finally, the estimator requires the initial state, which our experiments supply from the data. On a physical system without a reference sensor at deployment, it must be estimated separately. The model is also trained for a fixed time step $\Delta t$ and window length $o$, so irregularly sampled measurements or a different sampling rate require retraining or resampling.

To sum up, NormFlowSDE as a direct estimator performs strongly overall in accuracy, coverage, and latency on both the simulated and the real datasets, including the partially observed Lorenz--96 and the contact-rich robot dynamics. These results argue for amortized filtering wherever the dynamics are hard to model, the noise is not Gaussian, and inference must be fast and calibrated.

\bibliography{references}
\bibliographystyle{tmlr}
\clearpage
\appendix
\section{Weak convergence of the centered NormFlowSDE inference scheme}
\label{app:diffusion}\label{sec:weak_conv_norm_flowsde}

This appendix gives the full argument behind the continuous-time interpretation in Section~\ref{sec:method}. The starting point is the trained, finite-step transition. We first construct an inference scheme that preserves that transition at its training interval, then prove convergence of its trajectories as the inference step decreases. The result concerns a conditional generative model. It does not identify its coefficients with the physical dynamics of the observed system.

\subsection{What is held fixed, and why the noise must be centered}

Write $h_0=\Delta t_{\mathrm{train}}>0$ for the physical interval of the trained transition and $h=\Delta t_{\mathrm{inf}}>0$ for the inference step. Fix a finite time horizon $[0,T]$ and a deterministic, continuous schedule $\xi=(u(\cdot),c(\cdot))$ of controls and observation features. The fitted parameters, training step $h_0$, and schedule $\xi$ remain fixed throughout, and only the inference step $h$ tends to zero. Any step-size inputs to the networks are held at their training values. The initial state $z_0\in\R^d$ is supplied by the data and is also fixed. Every inference chain and the limiting process start from this same value, $Z_0^h=Z_0=z_0$. Conditioning on $z_0$ is implicit in all probabilities and expectations below. We retain the schedule in every coefficient, with
\begin{align*}
 f_\varphi(z,t;\xi)&:=f_\varphi(z,t,u(t),c(t)),\\
 g_\phi(\epsilon;z,t,\xi)&:=g_\phi(\epsilon;z,t,u(t),c(t)).
\end{align*}
The resulting coefficients must satisfy the stated assumptions for this fixed $\xi$, and the constants below may depend on $\xi$. At each trained time $t_k$, $c(t_k)$ is the feature computed from the preceding observation window defined in Section~\ref{sec:method}. 

We use $z\in\R^d$ for a fixed state value and $Z$ for a random state. At time $t$, the trained update is
\begin{equation}\label{eq:trainingupdate}
 \begin{aligned}
 Z'&=z+h_0 f_\varphi(z,t;\xi)+\sqrt{h_0}\,g_\phi(\epsilon;z,t,\xi),\\
 \epsilon\mid z,t,\xi&\sim\N(0,I_d).
 \end{aligned}
\end{equation}
where $f_\varphi$ is the drift network and $g_\phi$ is the conditional flow. Assume its conditional mean exists, and define
\begin{align}\label{eq:centerednoise}
 m_\phi(z,t;\xi)&=\E_\epsilon[g_\phi(\epsilon;z,t,\xi)],\\
 \eta_c(z,t,\epsilon;\xi)&=g_\phi(\epsilon;z,t,\xi)-m_\phi(z,t;\xi).\notag
\end{align}
Here and below, $\E_\epsilon$ denotes integration over $\epsilon\sim\N(0,I_d)$ with the displayed state, time, and schedule fixed. Thus $\E_\epsilon[\eta_c(z,t,\epsilon;\xi)]=0$. The effective drift of the trained transition is its expected increment per unit time:
\begin{equation}\label{eq:effectivedrift}
 b_{\mathrm{eff}}(z,t;\xi)
 =f_\varphi(z,t;\xi)+\frac{m_\phi(z,t;\xi)}{\sqrt{h_0}}
 =\frac{\E[Z'-z\mid z,t,\xi]}{h_0}.
\end{equation}
Using this drift and the centered noise gives the inference map 
\begin{equation}\label{eq:refinement}
 F_{h,t}(z,\epsilon;\xi)
 =z+h b_{\mathrm{eff}}(z,t;\xi)+\sqrt h\,\eta_c(z,t,\epsilon;\xi).
\end{equation}
We distinguish the fixed training step $h_0$ from the inference step $h$ to preserve the learned transition at $h=h_0$. Centering the stochastic increment and compensating the drift then avoid a divergent mean contribution as $h\to0$.

\subsection{The limiting process and sufficient assumptions}\label{sec:limit-statement}

Repeated application of the inference map in Equation~\eqref{eq:refinement}, at times $t_k=kh$, gives
\begin{equation}\label{eq:centered}
 Z^h_{k+1}=Z^h_k+h b_{\mathrm{eff}}(Z^h_k,t_k;\xi)
                 +\sqrt h\,\eta_c(Z^h_k,t_k,\epsilon_k;\xi).
\end{equation}
For a current state $z$, the increment has conditional mean $h b_{\mathrm{eff}}(z,t;\xi)$ and covariance $hA(z,t;\xi)$, where
\begin{equation}\label{eq:limitcovariance}
 A(z,t;\xi)=\E_\epsilon[\eta_c(z,t,\epsilon;\xi)\eta_c(z,t,\epsilon;\xi)^\top].
\end{equation}
This identifies the candidate diffusion: its drift is $b_{\mathrm{eff}}$ and its diffusion covariance is $A$. Write its noise coefficient as a matrix $\sigma$ satisfying $\sigma\sigma^\top=A$. The conditions below ensure that matching these local moments also leads to convergence of complete trajectories.

\begin{assumption}[Conditions for diffusion convergence]\label{ass:limit}
On the finite horizon $[0,T]$, assume the following.
\begin{enumerate}
\item \textnormal{\bfseries Regular drift and diffusion.}\label{ass:limit-regularity}
The functions $b_{\mathrm{eff}}(z,t;\xi)$ and a covariance factor $\sigma(z,t;\xi)$ with $\sigma\sigma^\top=A$ are jointly continuous in $(z,t)$, globally Lipschitz in $z$, and of at most linear growth in $z$. The Lipschitz and growth constants are uniform in $t\in[0,T]$. These conditions control the trajectories and ensure uniqueness of the limiting SDE.
\item \textnormal{\bfseries Controlled noise moments.}\label{ass:limit-moments}
The centered noise $\eta_c$ is jointly measurable, and its third moment is bounded on every bounded state region, uniformly in time:
\begin{equation}\label{eq:local-third-assumption}
 \sup_{\norm z\leq R,\,t\in[0,T]}
 \E_\epsilon\norm{\eta_c(z,t,\epsilon;\xi)}^3<\infty,
 \qquad R<\infty.
\end{equation}
This condition controls the Taylor approximation error and the probability of large increments.
\item \textnormal{\bfseries Given initial state and independent noise.}\label{ass:limit-initial}
Every chain starts at the data-provided state $Z_0^h=z_0$. With $z_0$ and $\xi$ fixed, the Gaussian inputs $\epsilon_k\sim\N(0,I_d)$ are independent across steps. Thus trajectory randomness comes from the generated increments, not from sampling an initial state.
\end{enumerate}
\end{assumption}

\begin{theorem}[Weak convergence of the centered inference scheme]\label{thm:weak}
Under Assumption~\ref{ass:limit}, the paths generated by Equation~\eqref{eq:centered}, held constant between grid points as $\bar Z^h_t=Z^h_{\lfloor t/h\rfloor}$, satisfy
\[
 \operatorname{Law}(\bar Z^h\mid\xi)
 \Longrightarrow\operatorname{Law}(Z\mid\xi)
 \qquad(h\to0),
\]
in $D([0,T],\R^d)$ with the Skorokhod $J_1$ topology. The limit $Z$ is the unique-in-law solution of
\begin{equation}\label{eq:limit}
 \begin{aligned}
 dZ_t&=b_{\mathrm{eff}}(Z_t,t;\xi)\,dt+\sigma(Z_t,t;\xi)\,dW_t,\\
 Z_0&=z_0.
 \end{aligned}
\end{equation}
Conditional on the fixed data, $W$ is a standard $d$-dimensional Brownian motion. For every three-times continuously differentiable test function $v$ with compact support, the discrete generators also converge uniformly on bounded state regions and $t\in[0,T]$, as specified in Equation~\eqref{eq:generator} below.
\end{theorem}

The theorem compares distributions of whole trajectories. The space $D([0,T],\R^d)$ contains right-continuous paths with left limits, including the piecewise-constant paths $\bar Z^h$. Its Skorokhod $J_1$ topology allows small shifts in jump times when comparing paths \citep{skorokhod1956limit}. The limiting SDE paths themselves are continuous. Section~\ref{sec:consistency-proof} now shows how the three assumptions turn the local moment match into this path-convergence result.

\subsection{Proof of Theorem~\ref{thm:weak}}\label{sec:consistency-proof}
The proof has two tasks: show that the trajectory distributions have limits, and identify those limits. Step~1 calculates the expected change over one inference step. Steps~2--4 then control whole trajectories: they prevent escape to arbitrarily large states, remove persistent jumps, and bound changes over short time intervals. Step~5 shows that any resulting limit must have the law of the SDE in Equation~\eqref{eq:limit}. We use the standard diffusion-approximation result of \citet{ethier1986markov,kushner2001numerical} to connect the path estimates and the limiting expected-change identity.

\paragraph{1. Identify the local diffusion limit.}
For a scalar test function $v\in C_c^3(\R^d)$, meaning that $v$ has three continuous derivatives and compact support, define
\begin{equation}\label{eq:discretegenerator}
 \mathcal L_{h,t}^{\xi}v(z)
 =\frac{\E_\epsilon[v(F_{h,t}(z,\epsilon;\xi))]-v(z)}h.
\end{equation}
This is the expected change of $v$ in one inference step, divided by the step duration. To calculate its limit, fix $(z,t,\xi)$ and write
$\Delta=h b_{\mathrm{eff}}(z,t;\xi)+\sqrt h\,\eta_c(z,t,\epsilon;\xi)$.
Throughout this local calculation, the arguments $(z,t,\xi)$ are held fixed and suppressed on $b_{\mathrm{eff}}$, $\eta_c$, and $A$:
\begin{align}\label{eq:proofmoments}
 \E[\Delta\mid z,t,\xi]
 &=h b_{\mathrm{eff}}+\sqrt h\,\E_\epsilon[\eta_c]
 =h b_{\mathrm{eff}},\notag\\
 \E[\Delta\Delta^\top\mid z,t,\xi]
 &=h^2b_{\mathrm{eff}}b_{\mathrm{eff}}^\top
   +h\E_\epsilon[\eta_c\eta_c^\top]+h^{3/2}\bigl(
 b_{\mathrm{eff}}\E_\epsilon[\eta_c]^\top
 +\E_\epsilon[\eta_c]b_{\mathrm{eff}}^\top\bigr)\notag\\
 &=h^2b_{\mathrm{eff}}b_{\mathrm{eff}}^\top+hA.
\end{align}
The $h^{3/2}$ line displays both drift--noise cross terms. They vanish because the noise is centered, $\E_\epsilon[\eta_c]=0$. The remaining noise term is its covariance $A$, defined in Equation~\eqref{eq:limitcovariance}.

Taylor's formula gives
\begin{equation}\label{eq:taylorexpansion}
 v(z+\Delta)-v(z)
 =\nabla v(z)^\top\Delta
  +\tfrac12\Delta^\top\nabla^2v(z)\Delta+R_v(z,\Delta),
 \qquad |R_v(z,\Delta)|\leq C_v\norm\Delta^3.
\end{equation}
Here $\nabla v$ and $\nabla^2v$ are the gradient and Hessian. To control the remainder, we need the \emph{third absolute moment of the full increment} $\Delta$. The drift is bounded on each region $\norm z\leq R$, $t\in[0,T]$, and the noise third moment is bounded by Equation~\eqref{eq:local-third-assumption}. Hence, for $0<h\leq1$,
\begin{align}\label{eq:thirdmoment}
 \E[\norm\Delta^3\mid z,t,\xi]
 &\leq4h^3\norm{b_{\mathrm{eff}}(z,t;\xi)}^3
   +4h^{3/2}\E_\epsilon\norm{\eta_c(z,t,\epsilon;\xi)}^3\notag\\
 &\leq C_{R,T}h^{3/2},
 \qquad \norm z\leq R,\quad t\in[0,T].
\end{align}
The constant is independent of $h$, $z$, and $t$ within this region. It is derived from, and distinct from, the assumed noise-moment bound.

Substituting the moments from Equation~\eqref{eq:proofmoments} into the expected Taylor expansion gives the full generator calculation:
\begin{align}\label{eq:generator-error}
 \mathcal L_{h,t}^{\xi}v(z)
 &=\frac1h\left(
 \nabla v(z)^\top\E_\epsilon\Delta
 +\frac12\operatorname{tr}\,\left(\nabla^2v(z)\E_\epsilon[\Delta\Delta^\top]\right)
 +\E_\epsilon R_v(z,\Delta)\right)\notag\\
 &=\frac1h\left(
 h\nabla v(z)^\top b_{\mathrm{eff}}
 +\frac12\operatorname{tr}\,\left(\nabla^2v(z)
       [hA+h^2b_{\mathrm{eff}}b_{\mathrm{eff}}^\top]\right)
 +\E_\epsilon R_v(z,\Delta)\right)\notag\\
 &=b_{\mathrm{eff}}^\top\nabla v(z)
   +\frac12\operatorname{tr}\,\left(A\nabla^2v(z)\right)
   +\frac h2 b_{\mathrm{eff}}^\top\nabla^2v(z)b_{\mathrm{eff}}
   +\frac{\E_\epsilon R_v(z,\Delta)}h .
\end{align}
The last two terms vanish uniformly on each bounded state region, since
\begin{align}\label{eq:generator-remainder}
 &\sup_{\norm z\leq R,\,t\in[0,T]}
 \left|\frac h2 b_{\mathrm{eff}}^\top\nabla^2v(z)b_{\mathrm{eff}}
       +\frac{\E_\epsilon R_v(z,\Delta)}h\right|\notag\\
 &\quad\leq
 \frac h2\sup_{\norm z\leq R,\,t\in[0,T]}
       \bigl(\norm{b_{\mathrm{eff}}}^2\norm{\nabla^2v(z)}_{\mathrm{op}}\bigr)
 +\frac{C_v}h\sup_{\norm z\leq R,\,t\in[0,T]}\E_\epsilon\norm\Delta^3
 \leq C_{R,T,v}(h+\sqrt h)\longrightarrow0.
\end{align}
Here $\norm{\cdot}_{\mathrm{op}}$ is the matrix operator norm. Thus the limiting generator is
\begin{equation}\label{eq:generator}
 \mathcal L_t^{\xi}v(z)
 =\lim_{h\to0}\mathcal L_{h,t}^{\xi}v(z)
 \overset{\mathrm{(\ref{eq:discretegenerator}),\,(\ref{eq:refinement})}}{=}
 \lim_{h\to0}\frac{
 \E_\epsilon[v(F_{h,t}(z,\epsilon;\xi))-v(z)]}{h}.
\end{equation}
Equations~\eqref{eq:generator-error}--\eqref{eq:generator-remainder} identify this limiting expected change with the drift and covariance terms of the candidate SDE. This proves the local conclusion of Theorem~\ref{thm:weak}. The remaining steps concern entire paths.

\paragraph{2. Bound the probability of leaving a large state region.}
The local calculation above is uniform only when the state stays in a bounded region. For a radius $R>\norm{z_0}$, we use the closed Euclidean ball centered at the origin,
\begin{equation}\label{eq:state-ball}
 \overline B(0,R):=\{z\in\R^d:\norm z\leq R\}.
\end{equation}
Here $\norm{\cdot}$ is the Euclidean norm, so the ball contains every state whose distance from the origin is at most $R$, including its boundary. We need to show that paths are unlikely to leave this ball over the entire time horizon, with a bound that does not deteriorate as $h$ decreases. We set $N_h=\lfloor T/h\rfloor$ and use the notation
\[
 b_k=b_{\mathrm{eff}}(Z_k^h,t_k;\xi),\qquad
 A_k=A(Z_k^h,t_k;\xi),\qquad
 \eta_k=\eta_c(Z_k^h,t_k,\epsilon_k;\xi).
\]
The linear-growth assumption gives
\begin{equation}\label{eq:innovation-growth}
 \norm{b_{\mathrm{eff}}(z,t;\xi)}^2+
 \E_\epsilon\norm{\eta_c(z,t,\epsilon;\xi)}^2
 =\norm{b_{\mathrm{eff}}(z,t;\xi)}^2+\operatorname{tr}A(z,t;\xi)
 \leq C(1+\norm z^2).
\end{equation}
We measure the size of a state by $V(z)=1+\norm z^2$. Centering and the growth bound imply, for $0<h\leq1$,
\begin{align}\label{eq:one-step-size}
 \E[V(Z_{k+1}^h)\mid Z^h_{0:k},\xi]
 &=1+\norm{Z_k^h+h b_k}^2+h\operatorname{tr}A_k\notag\\
 &\leq1+(1+h)\norm{Z_k^h}^2+(h+h^2)\norm{b_k}^2
          +h\operatorname{tr}A_k\notag\\
 &\leq(1+Ch)V(Z_k^h).
\end{align}
Here $Z^h_{0:k}$ denotes the generated history through step $k$. The bound says that the expected squared size grows by a factor at most $1+Ch$ per step.

To turn this into a bound for the whole path, we freeze a copy of the chain at its first exit from $\overline B(0,R)$. Exit from this closed ball means $\norm{Z_k^h}>R$:
\[
 \tau_R=\inf\{k\leq N_h:\norm{Z_k^h}>R\},\qquad
 Y_k^h=Z^h_{\min(k,\tau_R)},\qquad \inf\varnothing=\infty.
\]
This stopped copy is only a proof device and does not change the inference algorithm. Before exit it follows the original chain, including the step that crosses the boundary. After exit it stays at that exit state. Equation~\eqref{eq:one-step-size} therefore also bounds its expected size, and iteration gives
\begin{align}\label{eq:stopped-size}
 \E[V(Y_{k+1}^h)\mid Z^h_{0:k},\xi]
 &\leq(1+Ch)V(Y_k^h),\notag\\
 \E[V(Y_{N_h}^h)\mid\xi]
 &\leq(1+Ch)^{N_h}V(z_0)
 \leq e^{CT}(1+\norm{z_0}^2)=:C_T.
\end{align}
If any state leaves the ball, the frozen final state has squared norm greater than $R^2$. Hence
\begin{equation}\label{eq:containment}
 \Pr\,\left(\max_{k\leq N_h}\norm{Z_k^h}>R\,\middle|\,\xi\right)
 =\Pr(\tau_R\leq N_h\mid\xi)
 \leq\frac{\E[V(Y_{N_h}^h)\mid\xi]}{1+R^2}
 \leq\frac{C_T}{1+R^2}.
\end{equation}
The same bound applies to $\sup_{t\leq T}\norm{\bar Z_t^h}$. For any probability tolerance $\varepsilon>0$, we can choose $R$ large enough that $C_T/(1+R^2)\leq\varepsilon$. This single radius works for every $0<h\leq1$: the process $\bar Z^h$ remains in $\overline B(0,R)$ throughout $[0,T]$ with probability at least $1-\varepsilon$. Because closed bounded balls in $\R^d$ are compact, this is the \emph{compact containment condition} of \citet[Chapter~3, Remark~7.3, p.~129]{ethier1986markov}.

\paragraph{3. Combine the one-step and whole-path jump bounds.}
Fix a jump threshold $a>0$. Equation~\eqref{eq:thirdmoment} bounds the third absolute moment of the full increment. Since $\norm\Delta^2\leq\norm\Delta^3/a$ whenever $\norm\Delta>a$,
\begin{equation}\label{eq:smalljumps}
 \sup_{\norm z\leq R,\,t\in[0,T]}
 \frac1h\E[\norm\Delta^2\boldsymbol 1_{\{\norm\Delta>a\}}\mid z,t,\xi]
 \qquad\leq\frac1{ah}
 \sup_{\norm z\leq R,\,t\in[0,T]}\E[\norm\Delta^3\mid z,t,\xi]
 \leq\frac{C_{R,T}}a\sqrt h\longrightarrow0.
\end{equation}
The indicator selects increments larger than $a$. This proves that large increments contribute vanishing second moment per unit time.

We also need the probability of encountering even one large jump along the path. The same third-moment estimate bounds each such probability by $C_{R,T}h^{3/2}/a^3$ while the starting state is inside the ball. There are at most $T/h$ increments. Adding these probabilities and the escape probability from Step~2 gives
\begin{align}\label{eq:maximum-jump}
 \Pr\,\left(\max_{k<N_h}\norm{Z_{k+1}^h-Z_k^h}>a\,\middle|\,\xi\right)
 &\quad\leq\Pr(\tau_R\leq N_h\mid\xi)
 +\sum_{k<N_h}\Pr\,\left(\norm{Z_{k+1}^h-Z_k^h}>a,\quad k<\tau_R\,\middle|\,\xi\right)\notag\\
 &\quad\leq\frac{C_T}{1+R^2}+\frac{T}{h}\frac{C_{R,T}h^{3/2}}{a^3}
 =\frac{C_T}{1+R^2}+\frac{T C_{R,T}}{a^3}\sqrt h .
\end{align}
First choose $R$ large, then let $h$ decrease. The largest jump over the whole horizon tends to zero in probability.

\paragraph{4. Bound the change over a short time interval.}
Small individual jumps do not rule out many small increments adding up rapidly. For example, continuous paths $x_n(t)=\sin(nt)$ stay bounded and have no jumps, yet change by order one over intervals of length $1/n$. 

Start at a grid index $K\leq N_h$ that may be chosen from the generated history, such as the first threshold crossing, capped at $N_h$. The choice cannot use future noise: whether $K\leq k$ must be determined at step $k$. Look ahead a fixed number $m$ of steps, ending at $L=\min(K+m,N_h)$. Their duration is at most $\ell=mh$. We first bound the change of the stopped copy $Y^h$ from Step~2, so every active update starts inside the ball. Set
\[
 B_R=\sup_{\norm z\leq R,\,t\in[0,T]}\norm{b_{\mathrm{eff}}(z,t;\xi)},
 \qquad
 D_R=\sup_{\norm z\leq R,\,t\in[0,T]}\operatorname{tr}A(z,t;\xi).
\]
Both constants are finite. Let $I_j$ equal one when $K\leq j<L$ and $j<\tau_R$, and zero otherwise. There are at most $m$ active steps, so $\sum_j I_j\leq m$. The increment splits into accumulated drift and noise:
\begin{equation}\label{eq:interval-decomposition}
 Y_L^h-Y_K^h
 =\underbrace{\sum_{j<N_h}I_j h b_j}_{D_{K,m}^h}
  +\underbrace{\sum_{j<N_h}I_j\sqrt h\,\eta_j}_{M_{K,m}^h}.
\end{equation}
Each $I_j$ is determined before $\epsilon_j$ is drawn. Thus the active noise increments still have zero conditional mean, and cross terms between different noise increments vanish. Consequently,
\begin{align}\label{eq:interval-components}
 \norm{D_{K,m}^h}^2
 &\leq\left(\sum_{j<N_h}I_jh B_R\right)^2
 \leq B_R^2\ell^2,\notag\\
 \E[\norm{M_{K,m}^h}^2\mid\xi]
 &=\E\,\left[\sum_{j<N_h}I_jh\operatorname{tr}A_j\,\middle|\,\xi\right]
 \leq D_R\ell.
\end{align}
The drift grows with elapsed time, and the noise \emph{variance} also grows with elapsed time. We do not assume that the accumulated drift and noise are independent. Using $\norm{D+M}^2\leq2\norm D^2+2\norm M^2$ gives
\begin{equation}\label{eq:short-increment}
 \E[\norm{Y_L^h-Y_K^h}^2\mid\xi]
 \leq2B_R^2\ell^2+2D_R\ell.
\end{equation}
Except when the original path exits the ball, its increment equals that of the stopped copy. For $\ell\leq\delta+h$,
\begin{align}\label{eq:short-increment-probability}
 \Pr(\norm{Z_L^h-Z_K^h}>a\mid\xi)
 &\leq\Pr(\tau_R\leq N_h\mid\xi)
    +\frac{\E[\norm{Y_L^h-Y_K^h}^2\mid\xi]}{a^2}\notag\\
 &\leq\frac{C_T}{1+R^2}
   +\frac{2}{a^2}\left[B_R^2(\delta+h)^2+D_R(\delta+h)\right].
\end{align}
Choose $R$ large first. With that radius fixed, the second term tends to zero as the interval length $\delta$ and step $h$ decrease. Thus a large change is unlikely over a short interval, even when its starting time is selected from the generated history.

For starting times between grid points, rounding to the grid alters the compared increment by at most two grid jumps, and Step~3 shows that this error vanishes. The containment and short-interval estimates are the compactness conditions in the cited diffusion-approximation result: every sequence $h\to0$ has a subsequence whose path laws converge. This property is called \emph{tightness}. Together with the vanishing jumps, these bounds ensure that the possible limits have continuous paths.

\paragraph{5. Identify the limit through its expected changes.}
Steps~2--4 establish the existence of subsequential limits, but do not yet specify their law. Let $X$ denote any such limit. We now show why its law must be that of the SDE in Equation~\eqref{eq:limit}.

\emph{From one step to an interval.}
The generator predicts the expected change of any smooth test function $v$:
\begin{equation}\label{eq:conditional-test-increment}
 \E[v(Z^h_{j+1})-v(Z^h_j)-h\mathcal L_{h,t_j}^{\xi}v(Z_j^h)
       \mid Z^h_{0:j},\xi]=0.
\end{equation}
Thus \emph{actual change minus predicted change} has zero conditional mean. For grid indices $p<q$, sum these differences for the stopped copy from Step~2. The state changes telescope, giving
\begin{equation}\label{eq:stopped-expected-change}
 \E\,\left[
 \underbrace{v(Y_q^h)-v(Y_p^h)}_{\mbox{\scriptsize actual change}}
 -\underbrace{\sum_{j=p}^{q-1}\boldsymbol 1_{\{j<\tau_R\}}
       h\mathcal L_{h,t_j}^{\xi}v(Z_j^h)}_{\mbox{\scriptsize accumulated predicted change}}
 \,\middle|\,Z^h_{0:p},\xi\right]=0.
\end{equation}
Stopping ensures that every rate in this sum is evaluated inside the region where Step~1 gives uniform convergence.

\emph{From the discrete sum to a time integral.}
Replacing each discrete rate by $\mathcal L_{t_j}^{\xi}v$ incurs a total error bounded by
\[
 \left|\sum_{j=p}^{q-1}\boldsymbol 1_{\{j<\tau_R\}}h
  \bigl(\mathcal L_{h,t_j}^{\xi}v-\mathcal L_{t_j}^{\xi}v\bigr)(Z_j^h)\right|
 \leq T\sup_{\norm z\leq R,\,t\in[0,T]}
       |\mathcal L_{h,t}^{\xi}v(z)-\mathcal L_t^{\xi}v(z)|
 \longrightarrow0.
\]
Continuity of the limiting paths and coefficients turns the sum of limiting rates into its time integral. The cited diffusion-approximation result passes the stopped conditional-mean identity to the limit. Step~2 then allows $R$ to increase without retaining a stopped boundary. For $0\leq s\leq t\leq T$, the resulting identity is
\begin{equation}\label{eq:limit-expected-change}
 \E\,\left[
 \underbrace{v(X_t)-v(X_s)}_{\mbox{\scriptsize actual change}}
 -\underbrace{\int_s^t\mathcal L_r^\xi v(X_r)\,dr}_{\mbox{\scriptsize accumulated predicted change}}
 \,\middle|\,X_{[0,s]},\xi\right]=0,
 \qquad X_0=z_0.
\end{equation}
Here $X_{[0,s]}$ is the path observed up to time $s$. Both $v$ and $\mathcal L_t^\xi v$ are bounded because $v$ has compact support, so the limiting identity is valid over the full horizon.

\emph{Why this identifies the SDE.}
For the candidate SDE solution $Z$, It\^o's formula gives exactly the same rule:
\begin{align}\label{eq:ito-expected-change}
 v(Z_t)-v(Z_s)-\int_s^t\mathcal L_r^\xi v(Z_r)\,dr
 &=\int_s^t\nabla v(Z_r)^\top\sigma(Z_r,r;\xi)\,dW_r,\notag\\
 \E\,\left[\int_s^t\nabla v(Z_r)^\top\sigma(Z_r,r;\xi)\,dW_r
       \,\middle|\,Z_{[0,s]},\xi\right]&=0.
\end{align}
The stochastic integral represents the unpredicted fluctuation, and its conditional mean is zero. Requiring this expected-change rule for \emph{every} $v\in C_c^3$, together with continuous paths and $X_0=z_0$, is the standard characterization of the diffusion law used in the cited result. The regularity assumption guarantees that this law is unique. Hence
\[
 \mbox{any subsequential limit }X
 \quad\Longrightarrow\quad
 \operatorname{Law}(X\mid\xi)=\operatorname{Law}(Z\mid\xi).
\]
Every possible subsequential limit is therefore the same SDE law, so the entire family of path laws converges. This completes the proof of Theorem~\ref{thm:weak}.\hfill$\square$

\subsection{Accuracy of approximate centering}

The theorem uses the exact noise mean $m_\phi$. If it is replaced by an approximation $\widehat m_h$, the approximation affects both the drift compensation and the centered noise. The following result states how accurate it must be as the inference step decreases.

\begin{corollary}[Approximate conditional means]\label{cor:centering}
Suppose Assumption~\ref{ass:limit} holds. Replace $m_\phi$ by a deterministic measurable approximation $\widehat m_h$ in both the compensated drift and centered noise. If
\begin{equation}\label{eq:centererror}
 \sup_{z,\,t\in[0,T]}\norm{\widehat m_h(z,t;\xi)-m_\phi(z,t;\xi)}=o(\sqrt h),
\end{equation}
then the same diffusion limit holds.
\end{corollary}
\paragraph{Proof.}
Let $e_h(z,t;\xi)=\widehat m_h(z,t;\xi)-m_\phi(z,t;\xi)$ be the mean-estimation error. The drift increment gains $h e_h/\sqrt{h_0}$, while the noise increment loses $\sqrt h\,e_h$. Substituting the approximate mean and collecting these two changes gives
\begin{align}\label{eq:approximateupdate}
 \widehat Z'
 &=z+h\left(f_\varphi(z,t;\xi)+\frac{\widehat m_h(z,t;\xi)}{\sqrt{h_0}}\right)+\sqrt h\left(g_\phi(\epsilon;z,t,\xi)-\widehat m_h(z,t;\xi)\right)\notag\\
 &=F_{h,t}(z,\epsilon;\xi)
   +\left(\frac h{\sqrt{h_0}}-\sqrt h\right)e_h(z,t;\xi).
\end{align}
The last line separates the exact inference update from the error caused by approximate centering. For a continuous-time limit, the relevant error is the added drift \emph{per unit time}, obtained by dividing the extra term by $h$:
\begin{equation}\label{eq:centering-drift-error}
 \frac{\widehat Z'-F_{h,t}(z,\epsilon;\xi)}h
 =\left(\frac1{\sqrt{h_0}}-\frac1{\sqrt h}\right)e_h(z,t;\xi).
\end{equation}
Its norm is uniformly bounded by
\[
 \left(\frac1{\sqrt{h_0}}+\frac1{\sqrt h}\right)
 \sup_{z,\,t\in[0,T]}\norm{e_h(z,t;\xi)}\longrightarrow0
\]
under Equation~\eqref{eq:centererror}. Thus the added drift vanishes uniformly, while the centered random part remains unchanged. This bounded perturbation preserves the moment and path bounds in Steps~1--4, and the limiting expected-change rule in Step~5 is unchanged. The same diffusion limit follows.\hfill$\square$

At $h=h_0$, the error coefficient in Equation~\eqref{eq:approximateupdate} is zero: using the same mean estimate in the drift and noise preserves the trained update exactly. As $h\to0$, however, a fixed nonzero estimation error is amplified by the $1/\sqrt h$ factor in Equation~\eqref{eq:centering-drift-error}. 

\subsection{Scope of the convergence result}\label{sec:limit-scope}

Theorem~\ref{thm:weak} establishes convergence of conditional trajectory distributions for the centered inference scheme under the stated assumptions. Only the first two conditional moments remain in the limit, so non-Gaussian finite-step increments can approach an It\^o diffusion. The following distinctions specify what this result does and does not provide.

\paragraph{Weak convergence and individual paths.}
The theorem concerns distributions of paths. It does not compare an individual simulated path with an SDE path driven by the same Brownian realization, or prove that their samplewise error tends to zero. Such a strong-convergence statement requires a specified coupling and a separate error analysis.

\paragraph{The training step and smaller inference steps.}
The construction preserves one trained transition at $h=h_0$. This does not imply that several smaller steps reproduce that transition, or that the limiting SDE has the same transition law over an interval of length $h_0$. Convergence as $h\to0$ is distinct from equality of finite-interval distributions.

\paragraph{Numerical accuracy and learned models.}
The theorem gives no convergence rate or adaptive-step guarantee. Its regularity and moment assumptions must be checked for the learned coefficients. Finite-step sampling and density evaluation in Section~\ref{sec:method} remain valid without the optional centering used for this continuous-time argument.

\begin{figure}[t]
\centering\includegraphics[width=\linewidth]{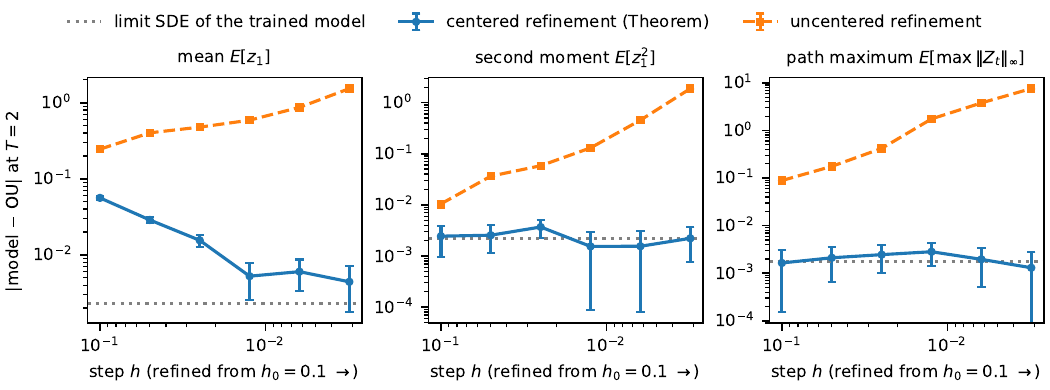}
\caption{Centered refinement of a trained NormFlowSDE transition recovers a standard SDE. The reference process is the two-dimensional Ornstein--Uhlenbeck SDE $dZ=-KZ\,dt+S\,dW$ with $K=\bigl(\begin{smallmatrix}1&0.8\\-0.8&1\end{smallmatrix}\bigr)$, $S=\bigl(\begin{smallmatrix}0.6&0\\0.3&0.4\end{smallmatrix}\bigr)$, and $Z_0=(2,-1)$. A NormFlowSDE model (drift network and conditional RealNVP noise flow, $\alpha=1.5$) was fitted by its exact one-step likelihood to transitions $z'=z-h_0Kz+\sqrt{h_0}\,S\eta$ at $h_0=0.1$, where $\eta$ has independent, skewed two-component-mixture coordinates with zero mean, unit variance and skewness $-0.38$. The finite-step increments are therefore non-Gaussian, while the only diffusion with the same first two moments is the reference process. With the fitted networks fixed, each panel shows the absolute error at $T=2$, relative to the exact Ornstein--Uhlenbeck law, of the mean of $z_1$, the second moment $z_1^2$, and the path maximum $\max_t\norm{Z_t}_\infty$ on the grid $t\in h_0\mathbb{N}$. The step decreases from $h_0$ to $h_0/32$ from left to right. Blue: the centered scheme of Theorem~\ref{thm:weak}, with $m_\phi$ and $A$ computed by $24\times24$ Gauss--Hermite quadrature. Orange: the uncentered refinement $z+hf_\varphi+\sqrt h\,g_\phi$. Dotted: the limit SDE of the fitted model, simulated by Euler--Maruyama at $h_0/128$. Its distance to the reference process is the learning error that remains as $h\to0$. The fitted drift $b_{\rm eff}$ and diffusion $A$ differ from $-Kz$ and $SS^\top$ by at most $0.027$ and $0.0045$ on $[-2,2]^2$. The centered error of the mean decreases from $0.056$ at $h_0$, the bias of the coarse training chain, towards this learning-error floor, whereas the other statistics are already near the floor at $h_0$. The fitted noise flow has a nonzero mean ($\norm{m_\phi}$ up to $0.83$), which is not identifiable from the drift in the trained update. Without centering, it acts as a drift of order $1/\sqrt h$, and the uncentered errors grow as $h$ decreases.}
\label{fig:weak-convergence-ou}
\end{figure}

\paragraph{Numerical illustration.}
Figure~\ref{fig:weak-convergence-ou} applies the centered inference scheme to a trained NormFlowSDE transition whose data-generating process is a standard SDE with a known solution. The trained update is held fixed, as in Theorem~\ref{thm:weak}, and only the inference step $h$ is reduced.

\section{Derivation of the reconstruction and transition losses}\label{app:training-loss}

All densities below condition on the supplied initial state $z_0$, observations, and controls. The sequential generator has normalized path density
\begin{equation}\label{eq:training-path-density}
 q_\omega(\mathbf z)
 =\prod_{k=0}^{N-1}q_\omega(z_{k+1}\mid z_k,t_k,u_k,c_k).
\end{equation}
The fixed-scale Gaussian reconstruction model and the resulting auxiliary joint density are
\begin{align}
 p_\sigma(x_k\mid z_k)&=\N(x_k;z_k,\sigma_{\mathrm{state}}^2I_d),
 \qquad \sigma_{\mathrm{state}}>0,\label{eq:reconstruction-model}\\
 p_{\omega,\sigma}(\mathbf x,\mathbf z)
 &=q_\omega(\mathbf z)\prod_{k=1}^N p_\sigma(x_k\mid z_k).\label{eq:training-joint-density}
\end{align}
Expanding the expected negative joint log density gives
\begin{align}\label{eq:joint-training-loss}
 \mathcal L_{\mathrm{joint}}(\omega;\mathbf x)
 &=\E_{\mathbf z\sim q_\omega}[-\log p_{\omega,\sigma}(\mathbf x,\mathbf z)]\notag\\
 &=\E_{\mathbf z\sim q_\omega}
 \left[-\sum_{k=1}^N\log p_\sigma(x_k\mid z_k)
       -\sum_{k=0}^{N-1}\log q_\omega(z_{k+1}\mid z_k,t_k,u_k,c_k)\right]\notag\\
 &=\mathcal L_{\mathrm{state}}(\omega;\mathbf x)+\mathcal L_{\mathrm{trans}}(\omega).
\end{align}
The Gaussian reconstruction density yields Equation~\eqref{eq:state-training-loss}. Its additive constant does not affect optimization at fixed $\sigma_{\mathrm{state}}$.

For a generated transition, the forward map from $\epsilon_k$ to $z_{k+1}$ has Jacobian $\sqrt{\Delta t}\,J_{\phi,k}(\epsilon_k)$. The change-of-variables identity in Equation~\eqref{eq:sampled-transition-density} therefore gives
\begin{align}\label{eq:self-transition-loss}
 \mathcal L_{\mathrm{trans}}(\omega)
 &=\E\left[\sum_{k=0}^{N-1}
 \left(-\log\N(\epsilon_k;0,I_d)+\log|\det J_{\phi,k}(\epsilon_k)|
       +\frac d2\log\Delta t\right)\right]\notag\\
 &=\E\left[\sum_{k=0}^{N-1}\log|\det J_{\phi,k}(\epsilon_k)|\right]
   +\frac{Nd}{2}\bigl[1+\log(2\pi\Delta t)\bigr].
\end{align}
The Gaussian term varies between sampled inputs, but its expectation is constant with respect to the trainable parameters. The remaining expected forward log determinant controls the path entropy.

\paragraph{Coupling-scale bound.}
Each affine coupling layer leaves one coordinate block unchanged and scales and shifts the other. Its log scales have the form $\alpha\tanh(\cdot)\in(-\alpha,\alpha)$, so each multiplicative scale lies in $(e^{-\alpha},e^\alpha)$. If layer $\ell$ transforms $d_\ell$ coordinates, its Jacobian is block triangular with an identity block and a diagonal block of those scales. Its log determinant is therefore the sum of the $d_\ell$ log scales and lies in $[-\alpha d_\ell,\alpha d_\ell]$. The shifts do not contribute to the determinant, and coordinate permutations have unit absolute determinant. Composing $L_{\mathrm f}$ layers gives $|\log|\det J_{\phi,k}(\epsilon)||\leq B=\alpha\sum_{\ell=1}^{L_{\mathrm f}} d_\ell$, uniformly over the input and conditioning variables, which is Equation~\eqref{eq:training-determinant-bound}. When the layers alternate between the two halves of an even-dimensional state, $\sum_\ell d_\ell=L_{\mathrm f}d/2$. The geometric mean of the singular values of $J_{\phi,k}$ then lies between $e^{-\alpha L_{\mathrm f}/2}$ and $e^{\alpha L_{\mathrm f}/2}$.

\paragraph{Entropy bound and noise floor.}
Equations~\eqref{eq:training-determinant-bound} and~\eqref{eq:self-transition-loss} imply
\begin{equation}\label{eq:transition-entropy-lower-bound}
 \mathcal L_{\mathrm{trans}}(\omega)
 \geq\frac{Nd}{2}\bigl[1+\log(2\pi\Delta t)\bigr]-NB.
\end{equation}
The entropy term is minimized by contraction, so wherever reconstruction does not oppose it, the bounded flow approaches this bound. The lower side of Equation~\eqref{eq:training-determinant-bound} then acts as a noise floor: the volume of the per-step increment $\sqrt{\Delta t}\,g_\phi$ cannot fall below $\Delta t^{d/2}e^{-B}$. The floor depends on $\alpha$ and $L_{\mathrm f}$ but not on the data. With two coupling layers and $\alpha=1$, for example, the geometric-mean noise scale cannot fall below $e^{-1}\approx0.37$.

\paragraph{Volume, not variance.}
Let $\sigma_1(\epsilon)\geq\cdots\geq\sigma_d(\epsilon)$ denote the singular values of $J_{\phi,k}(\epsilon)$. Equation~\eqref{eq:training-determinant-bound} constrains only their product, $|\det J_{\phi,k}(\epsilon)|=\prod_{i=1}^d\sigma_i(\epsilon)$, which is the local volume change of the map $\epsilon\mapsto g_\phi(\epsilon;z_k,t_k,u_k,c_k)$. It does not constrain the individual singular values. The bounded flow can therefore reduce its local scale below $e^{-B/d}$ in some directions only by enlarging it in others. A per-coordinate factor that adds $\sum_i\log s_{\phi,i}$ to the log determinant, and thereby lowers the attainable volume itself, is required. 

\paragraph{Learned noise scale.}
In Equation~\eqref{eq:learned-noise-scale}, the factor $s_{\phi,i}>0$ multiplies coordinate $i$ of the bounded flow output, so the standard deviation of that coordinate of the increment is scaled by $s_{\phi,i}$. It is a free parameter of the model and is constant along a trajectory. To keep it positive and finite, we restrict its logarithm to a fixed interval $[a_-,a_+]$ with $a_-<0<a_+$, using a sigmoid of an unconstrained parameter $r_{\phi,i}$:
\begin{equation}\label{eq:learned-scale-parameterization}
 \log s_{\phi,i}=a_-+(a_+-a_-)\,\mathrm{sigmoid}\bigl(r_{\phi,i}+r_0\bigr),
 \qquad r_0=\log(-a_-/a_+).
\end{equation}
The offset $r_0$ is chosen so that $r_{\phi,i}=0$ gives $s_{\phi,i}=1$, that is, no change to the flow. The parameters are initialized at zero. 

\paragraph{Effect on the bounds.}
Scaling coordinate $i$ by $s_{\phi,i}$ multiplies the volume by $s_{\phi,i}$, so the log determinant gains $\sum_i\log s_{\phi,i}$. This sum lies in $[d\,a_-,\,d\,a_+]$, and the bounded flow contributes a value in $[-B,B]$. Equation~\eqref{eq:training-determinant-bound} therefore still holds, with the lower exponent widened to $B_-=B+d\,|a_-|$ and the upper exponent to $B_+=B+d\,a_+$. Likewise, the entropy lower bound~\eqref{eq:transition-entropy-lower-bound} holds with $B$ replaced by $B_-$, which is $d\,|a_-|=10d$ lower per transition.

\paragraph{Why a proper score is needed.}
Because the entropy term rewards contraction, the much lower bound means that this term alone could shrink the noise almost to zero. The learned scale is therefore trained with an objective that also penalizes too little spread, such as the full-horizon energy score of Section~\ref{sec:training-rollout}. The energy score is a proper scoring rule: it is smallest when the generated spread matches the true conditional spread, whether that spread is large or small. With the bounded flow alone, a true spread below the floor cannot be reached. The learned scale makes it reachable.

Finally, the transition term equals $H(q_\omega)=\E_{q_\omega}[-\log q_\omega]$, giving
\begin{equation}\label{eq:joint-training-loss-entropy}
 \mathcal L_{\mathrm{joint}}(\omega;\mathbf x)
 =\mathcal L_{\mathrm{state}}(\omega;\mathbf x)+H(q_\omega).
\end{equation}
The marginal reference-state NLL, $-\log\int p_{\omega,\sigma}(\mathbf x,\mathbf z)\,d\mathbf z$, integrates out the generated path and is a different objective. If $q_\omega$ is used as the variational path density for the auxiliary model, the negative ELBO cancels the transition entropy:
\begin{equation}\label{eq:joint-versus-elbo}
 \begin{aligned}
 \E_{\mathbf z\sim q_\omega}
 \left[-\log p_{\omega,\sigma}(\mathbf x,\mathbf z)+\log q_\omega(\mathbf z)\right]
 &=\mathcal L_{\mathrm{state}}(\omega;\mathbf x)+H(q_\omega)-H(q_\omega)\\
 &=\mathcal L_{\mathrm{state}}(\omega;\mathbf x).
 \end{aligned}
\end{equation}
The proposed composite objective retains the entropy penalty.

\section{Implementation and reproducibility details}\label{app:reproducibility}
This appendix summarizes the data generation, the training of all methods, and the evaluation protocol needed to reproduce the results in Section~\ref{sec:experiments}.

\subsection{Data generation}\label{app:data}
Each dataset is a cache of fixed-length windows, each with a known initial state, a control sequence, noisy observations, and the reference states used for training and scoring. The data are split by trajectory or episode before windowing, and normalization statistics are computed on the training split only. Pendulum and Lorenz--96 metrics are reported in physical units. The observation noise of Equation~\eqref{eq:obsnoise} is drawn independently for every observed coordinate and time step. The caches are rebuilt deterministically from fixed seeds and were verified against stored checksums before training.

\paragraph{Pendulum.}
The dataset contains 320 trajectories of 60 steps that follow Equation~\eqref{eq:pendulum} with $g=9.81$, $\ell=m=1$, $c=0.2$, $\sigma_\theta=0.01$, and $\sigma_\omega=0.2$. They are integrated by Euler--Maruyama with eight substeps per interval $\Delta t=0.05$. The torque is a smooth, bounded random signal, $u_k=3\tanh(r_k)$, where $r_k$ is a first-order autoregressive process. The angle is unwrapped. Each trajectory is one window. The split is 224 / 48 / 48, and the noise scale is $\sigma=0.1$.

\paragraph{Lorenz--96.}
The dataset contains 2000 independent trajectories of Equation~\eqref{eq:lorenz96} ($d=12$, $F=8$, RK4 at $\Delta t=0.02$). Each trajectory starts from a small random perturbation of the origin and is integrated for 500 spin-up steps. The next 100 steps form the window, and the state at the end of the spin-up is the known initial state. Every second coordinate is observed with $\sigma=0.5$. The split is 1600 / 200 / 200 trajectories, of which the first 100 test windows are scored.

\paragraph{ALOHA insertion.}
The 50 teleoperation episodes of the simulated bimanual insertion task \citep{zhao2023aloha} at 50\,Hz are split 40 / 5 / 5. The 14 commanded joint positions are the controls and the 14 measured follower-joint positions, perturbed as described in Appendix~\ref{app:aloha-disturbance}, are the states. Windows of 250 steps are cut with a stride of 25 frames, giving 400 / 50 / 50 windows. States and controls are normalized per joint, and the noise ($\sigma=0.1$) is added in normalized units.

\paragraph{REASSEMBLE.}
The 148 episodes of the 30\,Hz real-robot dataset \citep{sliwowski2025reassemble} are split 94 / 17 / 37. The state is the seven joint positions and the nine-dimensional command is the control. Windows of 250 steps are cut without overlap. The training split is capped at 2000 windows, and 300 windows each are used for validation and test. Normalization and noise are as for ALOHA.

\subsection{ALOHA latent disturbance}\label{app:aloha-disturbance}
The recorded ALOHA states are deterministic given the commands, so we add a slowly varying latent disturbance per episode in joint-position units before normalization. With $q_k$ denoting the recorded joint positions and $u_k$ the commands, the disturbed joint positions $\tilde q_k$ and the disturbance $d_k$ follow
\begin{equation}\label{eq:alohadisturbance}
 \tilde q_{k+1}=\tilde q_k+(q_{k+1}-q_k)+\Delta t\,B\,d_k,\qquad
 d_{k+1}=\rho\,d_k+\sigma_d\,(1+\gamma\,r_k)\,\xi_k,\qquad \xi_k\sim\N(0,I_4),
\end{equation}
starting from $\tilde q_0=q_0$ and $d_0=0$, with $\rho=0.98$, $\sigma_d=0.05$, $\gamma=2$, $\Delta t=0.02$, and $B=0.35\,Q$ for a fixed random orthonormal $14\times4$ matrix $Q$. The regime variable $r_k$ is the command-tracking error of the frame relative to its median over the training episodes, capped at 3, so that the disturbance grows during contact. The disturbed trajectory $\tilde q$ is the reference state, and the commands are unchanged. The observation of $\tilde q_{k+1}$ adds the noise of Equation~\eqref{eq:obsnoise}.

\subsection{Evaluation protocol}\label{app:protocol}
All methods filter causally from the known initial state and never see future observations or intermediate reference states. For every test window a method returns $M$ posterior draws per time step ($M=128$, 200, 200, 500 on the pendulum, Lorenz--96, ALOHA, and REASSEMBLE). The particle filters run with the same particle count and return $M$ draws from the weighted particles. RMSE is computed per window and averaged over windows. CRPS is the empirical marginal CRPS of the draws, averaged over windows, steps, and coordinates. The energy score is computed on the state vector at each step. Coverage is the fraction of reference values inside the central interval between the empirical quantiles of the draws, pooled over windows, steps, and coordinates, and width is the mean length of that interval. Every method is trained with three seeds on the same caches and with the same total number of optimizer updates per dataset (8400, 18500, 9000, 11000, the largest of the four methods' tuned recipes). Checkpoints and hyperparameters are selected on the validation split, by the lowest validation RMSE unless stated otherwise, and no post-hoc calibration is applied. Tables report the mean and sample standard deviation over the three seeds. Latency is measured for the seed-0 checkpoint of every method on one NVIDIA L40S GPU with batch size one. Compilation is excluded, and the reported value is the median over 30 calls of the time to filter one test window.

\subsection{\method}\label{app:normflowsde}
\paragraph{Architecture.}
The feature network $T_\varphi$ of Equation~\eqref{eq:shared-feature} receives the state, the window feature, an embedding of the time, and, where available, the control. It is a multilayer perceptron with GELU activations, and its last hidden layer is the feature $\chi_k$. The drift is a linear layer on the feature. In each coupling layer of the conditional RealNVP noise flow, the log scales and shifts are computed from the unchanged noise coordinates and the feature by a network with one hidden tanh layer. The encoder is a bidirectional GRU that reads only the last $o=10$ measurements, each concatenated with its control (the window feature of Equation~\eqref{eq:sequence-conditioning}). Table~\ref{tab:nfsde-arch} lists the sizes, and Table~\ref{tab:params} lists the parameter counts of all methods.
\begin{table}[htbp]
\centering\footnotesize
\begin{tabular}{lcccc}
\toprule
 & Pendulum & Lorenz--96 & ALOHA & REASSEMBLE \\
\midrule
Window feature $d_c$ / encoder width & 16 / 16 & 128 / 128 & 64 / 64 & 64 / 64 \\
Feature network width / depth & 53 / 2 & 266 / 3 & 153 / 3 & 150 / 3 \\
Coupling layers / coupling hidden width & 2 / 16 & 2 / 64 & 2 / 64 & 2 / 64 \\
Log-scale bound $\alpha$ (stage 1 / 2) & 1 / 1 & 1.5 / 1.5 & 1 / 1 & 1 / 2 \\
\bottomrule
\end{tabular}
\caption{\method architecture per dataset. All four datasets use the explicit Euler update of Equation~\eqref{eq:normflowsde-overview} in training and inference.}
\label{tab:nfsde-arch}
\end{table}

\paragraph{Stage 1: composite objective.}
The first stage minimizes the objective of Equation~\eqref{eq:normflowsde_loss} on random training windows with the multiple-shooting partition of Section~\ref{sec:training}, using a bounded flow without the learned scale. Table~\ref{tab:nfsde-stage1} lists the dataset-specific settings. Four generated trajectories per window enter the state and transition terms. The rollout term is the energy score of Equation~\eqref{eq:energy-score-estimator} over the full window, evaluated every few updates with the stated number of draws. The solver uses the explicit Euler update of Equation~\eqref{eq:normflowsde-overview}. The optimizer is AdamW with gradient clipping. Lorenz--96 uses a warm-up and cosine learning-rate schedule, epoch-wise shuffling, and a small auxiliary one-step likelihood term, whereas the other datasets use a constant learning rate. The checkpoint with the lowest validation RMSE initializes the second stage.
\begin{table}[htbp]
\centering\footnotesize
\begin{tabular}{lcccc}
\toprule
 & Pendulum & Lorenz--96 & ALOHA & REASSEMBLE \\
\midrule
Updates & 4400 & 14500 & 5000 & 7000 \\
Trajectories per update & 16 & 64 & 1 & 4 \\
Learning rate & $6\times10^{-4}$ & $2\times10^{-3}$ & $3\times10^{-4}$ & $10^{-3}$ \\
Segment length (states) & 11 & 5 & 5 & 5 \\
$\sigma_{\mathrm{state}}$ (normalized units) & 1 & $1/\sqrt2$ & 1 & 1 \\
$\lambda_{\mathrm{cont}}$ & 1 & 0.1 & 1 & 1 \\
$\lambda_{\mathrm{rollout}}$ / rollout draws & 30 / 4 & 1 / 8 & 0 / -- & 0.5 / 4 \\
\bottomrule
\end{tabular}
\caption{Stage-1 settings of \method. On ALOHA the rollout term is switched off in stage 1 and enters only through stage 2.}
\label{tab:nfsde-stage1}
\end{table}

\paragraph{Stage 2: full-horizon energy score with the learned scale.}
The second stage warm-starts from the stage-1 checkpoint and adds the learned per-coordinate scale of Equation~\eqref{eq:learned-noise-scale}, initialized so that the model is unchanged at the first step. It minimizes the energy score of Equation~\eqref{eq:energy-score-estimator} of $M=16$ complete filtering rollouts per window, generated exactly as at inference, for 4000 updates (8000 on REASSEMBLE) on batches of 16 windows with a warm-up and cosine schedule (learning rates $3\times10^{-3}$, $3\times10^{-5}$, $3\times10^{-4}$, and $3\times10^{-5}$ on the four datasets). 

\subsection{Particle filters}\label{app:baselines}
The widths of all baselines were chosen so that each method's parameter count matches \method per dataset as closely as its architecture allows (Table~\ref{tab:params}).
\begin{table}[htbp]
\centering\footnotesize
\begin{tabular}{lcccc}
\toprule
Parameters ($\times10^3$) & Pendulum & Lorenz--96 & ALOHA & REASSEMBLE \\
\midrule
\method (stage 2) & 9.9 & 365.4 & 129.6 & 120.0 \\
NF-DPF & 9.6 & 366.3 & 128.9 & 120.0 \\
NSF-PF & 9.9 & 365.1 & 129.6 & 119.8 \\
FlowSDE-PF & 9.9 & 365.7 & 129.1 & 120.4 \\
\method-PF & 9.8 & 364.7 & 129.4 & 119.8 \\
Latent NSF & 9.7 & 364.5 & 130.0 & 120.2 \\
Latent SDE & 10.0 & 366.0 & 129.1 & 120.3 \\
SigKer SDE & 9.8 & 366.2 & 130.0 & 120.1 \\
\bottomrule
\end{tabular}
\caption{Parameter counts of the selected models. The count of \method includes the $d$ parameters of the learned scale, which the stage-1 model does not have. The \method-PF model is a transition of the same architecture with the learned scale plus its own measurement flow with the architecture of NSF-PF.}
\label{tab:params}
\end{table}

\paragraph{NF-DPF.}
Our implementation of \citet{chen2025flowparticle} has a conditional-RealNVP transition, an observation-conditioned proposal flow, and a conditional-RealNVP measurement density, with importance weights that include the transition and proposal densities. Training has two stages. The first trains on short 8-step windows with 32 particles and differentiable optimal-transport resampling, starting with the supervised likelihoods of the three components and then switching to the end-to-end filtering error. The second warm-starts the best stage-1 checkpoint and trains on a curriculum of increasing horizons up to the full window with 64 particles, minimizing the weighted-particle CRPS together with the measurement likelihood. At test time the filter runs with the protocol particle count and systematic resampling.

\paragraph{NSF-PF and FlowSDE-PF.}
The transition is the exact-density Neural Stochastic Flow of \citet{kiyohara2025nsf} or a FlowSDE drift \citep{lou2026flowsde} with fixed isotropic diffusion trained by bridge flow matching and integrated by Euler--Maruyama with four substeps per interval. Both transitions condition on the control and the time but not on observations. The measurement density $q(y\mid x)$ is a conditional RealNVP with four coupling layers (two on the pendulum), and all four particle filters use this architecture. Each filter trains its own measurement flow, and no trained network is shared between filters. Dynamics and measurement flow are trained jointly by maximum likelihood on random training windows for the full update budget. The filter is a bootstrap filter: particles are propagated by the transition without access to the measurement, weighted by $q(y\mid x)$, and resampled systematically when the effective sample size falls below half the particle count. During training, a bootstrap particle filter is run on the validation windows at regular intervals. Checkpoints and a process-noise inflation are selected jointly by validation RMSE, which is an isotropic Gaussian jitter added to every predicted particle.

\paragraph{\method-PF.}\label{app:normflowsde-pf}
The transition has the architecture of \method (Appendix~\ref{app:normflowsde}): a feature network with a linear drift head and the conditional RealNVP noise flow with the learned per-coordinate scale of Equation~\eqref{eq:learned-noise-scale}. Its feature network has widths 63, 363, 165, and 169 on the four datasets, chosen so that the filter with its measurement flow matches the parameter count of \method (Table~\ref{tab:params}). It conditions on the control and the time but not on observations, and it uses the explicit Euler update, for which the change-of-variables density is exact. The measurement density, the joint maximum-likelihood training, the update budget, and the joint validation selection of the checkpoint and the process-noise inflation are those of NSF-PF.

Table~\ref{tab:pf-hparams} lists the hyperparameters of the final particle filters.

\begin{table}[htbp]
\centering\footnotesize
\resizebox{\linewidth}{!}{%
\begin{tabular}{llcccc}
\toprule
Method & Hyperparameter & Pendulum & Lorenz--96 & ALOHA & REASSEMBLE \\
\midrule
All four & Transition hidden layers & 2 & 3 & 3 & 3 \\
 & Measurement flow: coupling layers & 2 & 4 & 4 & 4 \\
 & Measurement flow: hidden width / layers & 16 / 2 & 64 / 2 & 64 / 2 & 64 / 2 \\
\midrule
NF-DPF & Hidden width & 32 & 155 & 70 & 77 \\
 & Coupling layers & 2 & 2 & 2 & 2 \\
 & Stage 1: updates & 4400 & 12500 & 3000 & 5000 \\
 & Stage 1: particles & 32 & 32 & 32 & 32 \\
 & Stage 2: updates & 4000 & 6000 & 6000 & 6000 \\
 & Stage 2: learning rate & $10^{-4}$ & $10^{-4}$ & $10^{-4}$ & $10^{-4}$ \\
 & Stage 2: horizons & 8, 25, 60 & 8, 25, 50, 100 & 8, 25, 50, 250 & 8, 25, 50, 250 \\
 & Stage 2: windows per update & 32, 16, 8 & 32, 16, 8, 8 & 32, 16, 8, 8 & 32, 16, 8, 8 \\
 & Stage 2: particles & 64 & 64 & 64 & 64 \\
\midrule
NSF-PF, FlowSDE-PF, & Optimizer updates & 8400 & 18500 & 9000 & 11000 \\
\method-PF & Batch size (windows) & 16 & 16 & 16 & 16 \\
 & Learning rate & $3\times10^{-4}$ & $3\times10^{-4}$ & $3\times10^{-4}$ & $3\times10^{-4}$ \\
\midrule
NSF-PF & Hidden width & 48 & 224 & 107 & 113 \\
 & Coupling layers & 2 & 2 & 2 & 2 \\
 & Inflation $\tau$ & 0 & 0.005 & 0.05 & 0.03 \\
\midrule
FlowSDE-PF & Hidden width & 59 & 282 & 141 & 144 \\
 & Diffusion coefficient & 0.01 & 0.01 & 0.05 & 0.03 \\
 & Euler--Maruyama substeps & 4 & 4 & 4 & 4 \\
 & Inflation $\tau$ & 0.02, 0.03, 0.01 & 0 & 0.005 & 0 \\
\midrule
\method-PF & Feature network width & 63 & 363 & 165 & 169 \\
 & Coupling layers / coupling hidden width & 2 / 16 & 2 / 64 & 2 / 64 & 2 / 64 \\
 & Log-scale bound $\alpha$ & 1 & 1.5 & 1 & 1 \\
 & Inflation $\tau$ & 0 & 0.005 & 0.05 & 0.03 \\
\bottomrule
\end{tabular}
}
\caption{Hyperparameters of the final particle filters. The process-noise inflation $\tau$ is the standard deviation of the Gaussian jitter added to every predicted particle, selected per seed on the validation windows from $\{0, 0.005, 0.01, 0.02, 0.03, 0.05, 0.075, 0.1, 0.15, 0.2\}$. The three seeds selected the same value except for FlowSDE-PF on the pendulum, where the values of seeds 0, 1, and 2 are listed. All other entries are identical for the three seeds.}
\label{tab:pf-hparams}
\end{table}

\subsection{Direct baselines: latent NSF, latent SDE, and signature-kernel SDE}\label{app:direct-baselines}
The three direct baselines use the encoder of \method (Equation~\eqref{eq:sequence-conditioning}) with its settings: a bidirectional GRU of the width of \method (16, 128, 64, and 64 on the pendulum, Lorenz--96, ALOHA, and REASSEMBLE) over the window of the $o=10$ most recent measurements, in which the known initial state, restricted to its observed coordinates, takes the place of the measurement at $k=0$.

\paragraph{Latent NSF.} The prior transition is an NSF with the depth and coupling layers of NSF-PF, and the emission model is the measurement flow of the particle filters. The published posterior of Equation~\eqref{eq:latent-nsf-posterior} never sees the known initial state, which leaves the unobserved Lorenz--96 coordinates unidentifiable. We therefore make the posterior Markov in the state, as in structured inference networks \citep{krishnan2017structured}: $q_\varphi(x_{k+1}\mid x_k,y_{0:k})=\N(x_k+m_k,\mathrm{diag}\,s_k^2)$. Here $(m_k,s_k)$ are computed from $x_k$, the GRU state $h_k$, the control, and the time by a network with the width and depth of the prior. The posterior is rolled out from the known initial state like the other direct estimators. The loss is the $\beta$-NELBO of \citet[Equation~16]{kiyohara2025nsf}, with the KL terms estimated by Monte Carlo along an ancestral posterior path, a KL weight $\beta$ selected below, and linear annealing over the first 30\% of the updates. State supervision adds the negative log density of each true state under the posterior along that path and the NSF negative log likelihood of the true states at gaps of one to four steps. For gaps longer than one step, the prior is conditioned on the mean control over the gap. Because every gap is supervised directly, the skip-ahead KL, which has no closed form for a Markov posterior, and the flow consistency penalty are omitted. Samples are posterior paths from the known initial state. With the selected $\beta=0$, the KL terms vanish and the NSF prior no longer shapes the posterior, which is then trained by the state supervision alone.

\paragraph{Latent SDE.} Following \citet{li2020scalable}, the prior SDE $dx=f_\theta(x,u,t)\,dt+g_\theta(x,u,t)\,dW$ and the posterior SDE $dx=h_\varphi(x,h_k,u,t)\,dt+g_\theta(x,u,t)\,dW$ share a diagonal diffusion. Both are discretized by Euler--Maruyama with one step per interval. The ELBO is the Gaussian log likelihood of the true states under the posterior path, with a learned per-coordinate decoder standard deviation, minus the path KL divergence $\tfrac12\sum\|(h_\varphi-f_\theta)/g_\theta\|^2\Delta t$, with the KL weight annealed as above. Samples are posterior paths plus decoder noise.

\paragraph{Signature-kernel SDE.} Following \citet{issa2023sigkernel}, the generator is an SDE whose drift and diagonal diffusion receive the state, $h_k$, the control, and the time, integrated from the known initial state by Euler--Maruyama. For each training window, $m=8$ generator paths are compared with the reference state path by the unbiased estimate of the score in Equation~\eqref{eq:sigker-score}. Every path includes the initial state and is augmented with normalized time. The signature kernel uses an RBF static kernel in the normalized state coordinates, whose bandwidth (default $\sqrt d$) is selected below, and it is computed with the explicit second-order Goursat scheme of \citet{salvi2021goursat}. The training is non-adversarial, since there is no discriminator. This baseline replaces an SDE-GAN \citep{kidger2021gan} with the same generator and a neural-CDE discriminator on the joint path $(t,x,y,u)$, whose conditional Wasserstein training with gradient penalty collapsed to near-zero spread on every dataset.

All three use the recipe of NSF-PF: batches of 16 random training windows, AdamW with warm-up and cosine decay, gradient clipping at 1, and the update budget of the dataset (8400, 18500, 9000, 11000). The checkpoint with the lowest validation RMSE, computed from 64 draws on all validation windows at the evaluation interval of NSF-PF, is tested with the protocol's draw count. Widths were chosen to match the parameter count of \method (Table~\ref{tab:params}). Table~\ref{tab:baseline-hparams} lists the hyperparameters of the final models. The learning rate, the KL weight, and the kernel bandwidth were selected per dataset on seed 0 by the same validation RMSE, from the learning rates $\{3\times10^{-4},10^{-3}\}$, the KL weights $\beta\in\{1,0.1,0.01,0\}$ for latent NSF and $\beta\in\{1,0.1,0.01\}$ for the latent SDE, and the RBF bandwidths $\{\sqrt d,1,0.5\}$, and the selected configuration was then trained with two further seeds. Several selections lie at the edge of this grid, and many selected checkpoints are among the last of the budget, so the baselines might improve further with larger learning rates or budgets.

\begin{table}[htbp]
\centering\footnotesize
\begin{tabular}{llcccc}
\toprule
Method & Hyperparameter & Pendulum & Lorenz--96 & ALOHA & REASSEMBLE \\
\midrule
All three & Optimizer updates & 8400 & 18500 & 9000 & 11000 \\
 & Batch size (windows) & 16 & 16 & 16 & 16 \\
 & Encoder width & 16 & 128 & 64 & 64 \\
 & Hidden layers & 2 & 3 & 3 & 3 \\
\midrule
Latent NSF & Learning rate & $10^{-3}$ & $10^{-3}$ & $3\times10^{-4}$ & $10^{-3}$ \\
 & KL weight $\beta$ & 1 & 0 & 0 & 1 \\
 & Hidden width & 32 & 138 & 63 & 70 \\
\midrule
Latent SDE & Learning rate & $10^{-3}$ & $10^{-3}$ & $10^{-3}$ & $10^{-3}$ \\
 & KL weight $\beta$ & 0.1 & 0.01 & 0.01 & 0.01 \\
 & Hidden width & 44 & 178 & 107 & 107 \\
\midrule
SigKer SDE & Learning rate & $10^{-3}$ & $10^{-3}$ & $10^{-3}$ & $10^{-3}$ \\
 & RBF bandwidth & 0.5 & $\sqrt d$ & $\sqrt d$ & 1 \\
 & Hidden width & 50 & 203 & 126 & 124 \\
\bottomrule
\end{tabular}
\caption{Hyperparameters of the final direct baselines, identical for the three seeds of each cell. The encoder window is ten measurements on every dataset, as for \method.}
\label{tab:baseline-hparams}
\end{table}

\subsection{Model-based reference filters}\label{app:references}
On the pendulum and Lorenz--96 the data-generating model is known. The EKF and UKF use the true dynamics with the moment-matched Gaussian likelihood of the mixture. The bootstrap particle filter uses the true dynamics with the exact mixture likelihood and the protocol particle count. The process-noise scale of each filter is selected on the validation windows from a small grid, and the smallest grid value was selected for every filter and seed. On the pendulum this value is the true process noise. Lorenz--96 is deterministic, so a small additive state noise keeps the filters from collapsing. Three filter seeds are run on the same caches as the learned methods. Table~\ref{tab:reference-hparams} lists the settings.

\begin{table}[htbp]
\centering\footnotesize
\begin{tabular}{llcc}
\toprule
Filter & Setting & Pendulum & Lorenz--96 \\
\midrule
All three & Dynamics & Euler, 8 substeps & Runge--Kutta 4 \\
 & Process noise & $\sqrt{\kappa}\,(\sigma_\theta,\sigma_\omega)$, $\kappa=1$ & additive, std.\ 0.001 per step \\
 & Process-noise grid & $\kappa\in\{1,2,4,8\}$ & $\{0.001,0.003,0.01,0.03,0.1,0.3\}$ \\
 & Posterior draws $M$ & 128 & 200 \\
\midrule
EKF & Linearization & exact Jacobian & exact Jacobian \\
\midrule
UKF & Sigma points & $2d=4$, equal weights & $2d=24$, equal weights \\
\midrule
Bootstrap PF & Particles & 500 & 1000 \\
 & Resampling & systematic, if $\mathrm{ESS}<N/2$ & systematic, if $\mathrm{ESS}<N/2$ \\
\bottomrule
\end{tabular}
\caption{Settings of the model-based reference filters. The dynamics are those of the data generation (Appendix~\ref{app:data}). The EKF and UKF return $M$ draws from their Gaussian filtering distributions, and the particle filter returns $M$ draws from its weighted particles. $d$ is the state dimension and $N$ the particle count. $\mathrm{ESS}$ is the effective sample size.}
\label{tab:reference-hparams}
\end{table}

\subsection{Software, hardware, and code availability}\label{app:scripts}
All models are implemented in JAX 0.10 with Equinox 0.13 and Optax 0.2 under Python 3.12. Training ran on single NVIDIA A30 GPUs or on CPU nodes, and all latencies were measured on one NVIDIA L40S GPU (Appendix~\ref{app:protocol}). The code, the manuscript, and the frozen run directories are available at \url{https://github.com/papersubmit-a/NormFlowSDE}. The layout of the archive and the commands that rebuild the tables and figures are documented in its \path{README.md}. The code is released under the MIT License. The posterior and data arrays are not included in the repository because of their size and redistribution restrictions.

\end{document}